\documentclass[12pt]{article}
\usepackage[left=1.25in,top=1.0in,right=1.25in,bottom=1.0in]{geometry}
\usepackage{amsmath,amssymb,amsthm,amsfonts}
\usepackage[pdftex]{graphicx}
\usepackage{mathtools}
\usepackage{subcaption}
\usepackage{suffix}
\usepackage{color}
\usepackage{paralist}

\usepackage{datetime}

\usepackage{lscape}
\usepackage{longtable}
\usepackage{tabularx}

\RequirePackage[colorlinks,citecolor=blue,urlcolor=blue]{hyperref}
\usepackage[authoryear]{natbib}

\newdateformat{monthyear}{\monthname[\THEMONTH] \THEYEAR}

\allowdisplaybreaks

\usepackage{xifthen}

\newcommand{\diag}{{\rm diag}}

\renewcommand{\Re}{{\mathbb{R}}}

\newcommand{\ben}{\begin{enumerate}}
\newcommand{\een}{\end{enumerate}}
\newcommand{\beq}{\begin{equation}}
\newcommand{\eeq}{\end{equation}}

\newcommand{\half}{\frac{1}{2}}

\newcommand{\argmin}{\operatornamewithlimits{argmin}}

\newcommand{\N}{\mathrm{N}}

\newcommand*{\prox}{
  \ensuremath{
    \operatorname*{prox}
  }
}
\newcommand*{\moreau}[2]{
  \ensuremath{
    {#2}^{\ifthenelse{\equal{#1}{}}{1}{#1}}
  }
}

\newcommand{\sgn}{\mathop{\mathrm{sgn}}}

\newcommand*{\vnorm}[1]{\left|\left|#1\right|\right|}
\newcommand*{\enorm}[1]{\Vert #1 \Vert_2}

\newcommand{\bx}{{\bf x}}

\newcommand{\defeq}{\operatorname{:=}}

\newtheorem{Exa}{Example}

\graphicspath{{../../figures/}{../figures/}{./figures/}{./}}

\title{Proximal Algorithms in Statistics and Machine Learning}
\author{
  Nicholas G. Polson\\
  \textit{Booth School of Business}\\
  \textit{University of Chicago}
  \footnote{
    Professor of Econometrics and Statistics
    at the Chicago Booth School of Business. 
    email: ngp@chicagobooth.edu. 
  }
\and
  James G. Scott\\
  \textit{McCombs School of Business}\\
  \textit{University of Texas at Austin}
  \footnote{
    Associate Professor of Statistics at the McCombs School of Business, University
    of Texas at Austin.  
    email: James.Scott@mccombs.utexas.edu. 
  }
\and
  Brandon T. Willard\\
  \textit{Booth School of Business}\\
  \textit{University of Chicago}
  \footnote{ 
    bwillard@uchicago.edu 
    \newline
    We thank the participants at the 2014 ASA meetings for their comments.  
  }
}
\date{First Draft: October 2014\\
This Draft: \monthyear\today{}}

\begin{document}

\maketitle
\begin{abstract}

  \noindent In this paper we develop proximal methods for statistical learning.
  Proximal point algorithms are useful in statistics and
  machine learning for obtaining optimization solutions for composite functions.
  Our approach exploits closed-form solutions of proximal operators and
  envelope representations based on the Moreau,
  Forward-Backward, Douglas-Rachford and Half-Quadratic envelopes. Envelope
  representations lead to novel proximal algorithms for statistical
  optimisation of composite objective functions which include both non-smooth and
  non-convex objectives.  We illustrate our methodology with
  regularized Logistic and Poisson regression and non-convex bridge penalties
  with a fused lasso norm.  We provide a discussion of convergence of
  non-descent algorithms with acceleration and for non-convex functions.  
  Finally, we provide directions for future research.

  \vspace{0.1in}
\end{abstract}

\newpage


\section{Introduction}

  Our goal is to introduce statisticians to the large body of
  literature on \textit{proximal algorithms} for solving 
  optimization problems that arise within statistics.  
  By a proximal algorithm, we mean
  an algorithm whose steps involve evaluating the \textit{proximal operator} of
  some term in the objective function.  Both of these concepts will be defined
  precisely in the next section.  The canonical optimization problem 
  of minimising a measure of fit, together with a regularization penalty, sits
  at the heart of modern statistical practice and it arises, for example, in
  sparse regression \citep{tibshirani1996regression}, spatial smoothing
  \citep{tibs:fusedlasso:2005}, covariance estimation
  \citep{witten:tibs:hastie:2009}, image processing
  \citep{geman1992constrained,geman1995nonlinear,rudin:osher:faterni:1992},
  nonlinear curve fitting \citep{tibs:2014a}, Bayesian MAP inference
  \citep{Polson:Scott:2010b}, multiple hypothesis testing
  \citep{tansey:etal:2014} and shrinkage/sparsity-inducing prior regularisation
  problems \citep{2015arXiv150201148G}. For recent surveys on proximal
  algorithms, see 
  \citep{cevher2014convex, komodakis2014playing, combettes2011proximal,boyd2011distributed}.

  The techniques we employ here are often referred to as Proximal
  Gradient, Proximal Point, Alternating Direction Method of Multipliers (ADMM)
  \citep{boyd2011distributed}, Divide and Concur (DC), Frank-Wolfe (FW), Douglas-Ratchford (DR)
  splitting or alternating split Bregman (ASB) methods.  The
  field of image processing has developed many of these ideas in the form of Total
  Variation (TV) de-noising and half-quadratic (HQ) optimization
  \citep{geman1995nonlinear,geman1992constrained,nikolova2005analysis}.  
  Other methods such as fast iterative shrinkage
  thresholding algorithm (FISTA), expectation maximization (EM),
  majorisation-minimisation (MM) and iteratively reweighed least squares (IRLS)
  fall into our proximal framework.  Although such approaches are commonplace in 
  statistics and machine learning \citep{bien2013lasso}, there hasn't
  been a real focus on the general family of approaches that underly these
  algorithms. Early work on iterative proximal fixed point algorithms in Banach
  spaces is due to \citep{von1951functional, bregman1967relaxation,
  hestenes1969multiplier, martinet1970breve,rockafellar1976monotone}.  

  A useful feature of proximal algorithms are acceleration techniques
  \citep{nesterov1983method} which lead to non-descent algorithms that can provide an
  order-of-magnitude increase in efficiency.  When both functions are convex, and
  one has a smooth Lipschitz continuous gradient, a simple convergence result
  based on the reverse Pythagoras inequality is available.  Convergence rates of
  the associated gradient descent algorithms can vary and typically each analysis
  has to be dealt with on a case-by-case basis.  We illustrate acceleration for a
  sparse logistic regression with a fused lasso penalty.

  The rest of the paper proceeds as follows.  Section~\ref{sec:preliminaries}
  provides notation and basic properties of proximal operators and envelopes.
  Section~\ref{sec:prox_operators} describes the proximal operator and Moreau envelope.
  Section~\ref{sec:basic_prox_algos} describes the basic proximal algorithms and
  their extensions.  Section~\ref{sec:related_algos} describes common algorithms and techniques, such as ADMM and Divide and Concur, that rely on proximal algorithms.
  Section~\ref{sec:envelopes} discusses envelopes and how proximal algorithms
  can be viewed as envelope gradients. 
  Section~\ref{sec:prox_algos_composite} considers the general problem of composite
  operator optimisation and shows how to compute the exact proximal operator
  with a general quadratic envelope and a composite 
  regularisation penalty.  Section~\ref{sec:applications} illustrates the
  methodology with applications to logistic and Poisson regression with fused
  lasso penalties.  A bridge regression penalty illustrates the non-convex case and
  we apply our algorithm to the prostate data of \citet{hastie2009elements}.

  Table~\ref{tab:prox} provides commonly used proximal operators,
  Table~\ref{tab:hq} documents examples of half-quadratic envelopes and
  Table~\ref{tab:rates} lists convergence rates for a variety of algorithms.
  Appendix~\ref{app:convergence} discusses convergence results for both convex
  and non-convex cases together with Nesterov acceleration. Finally,
  Section~\ref{sec:discussion} concludes with directions for future research.
 
  \subsection{Preliminaries}
  \label{sec:preliminaries}

  Many optimization problems in statistics take the following form
  \begin{equation}
    \argmin_{x \in \mathcal{X}}  F(x)  \defeq l(x) + \phi(x) 
    \label{eq:general_objective}
  \end{equation} 
  where $l(x)$ is a measure of fit depending implicitly on some observed data
  $y$, $\phi(x)$ is a regularization term that imposes structure or effects
  a favorable bias-variance trade-off. Typically, $l(x)$ is a smooth function
  and $\phi(x)$ is non-smooth--like a lasso or bridge penalty--so as to induce
  sparsity. We will assume that $l$ and $\phi$ are convex and lower semi-continuous except when explicitly stated to be non-convex.
  
  We use $x=(x_1, \ldots x_d)$ to denote a $d$-dimensional parameter of interest, $y$ an
  $n$-vector of outcomes, $A$ a fixed $n \times d$ matrix whose rows are
  covariates (or features) $a_i^T$, and $B$ a fixed $k \times d$ matrix to
  encode some structural penalty on the parameter (as in the group lasso or
  fused lasso), $b$ are prior loadings and centerings and $\gamma>0$ is a
  regularisation parameter that will trace out a solution path.  All together,
  we have a composite objective of the form 
  \begin{equation}
    F(x) \defeq \sum_{i=1}^n l( y_i , a_i^T x ) + 
    \gamma \sum_{j=1}^k \phi \left ( [ B x - b ]_j \right )
   \label{eqn:canonicalform}
  \end{equation}
  For example, lasso can be viewed as a simple statistical model with the
  negative log likelihood
  from $y = Ax + \epsilon$, where $\epsilon$ is a standard normal measurement
  error, corresponding to the norm 
  $l(x) = \|A x - y\|^2$, and each parameter $x_j$ has independent Laplace priors corresponding to the regularisation penalty
  $\phi(x) = \gamma \sum_{j=1}^d |x_j|$.  

  Throughout, observations will be indexed by $i$, parameters by $j$, and iterations of an
  algorithm by $t$.  Unless stated otherwise, all functions are lower
  semi-continuous and convex (e.g. $l(x)$, $\phi(x)$), and all vectors are
  column vectors.  We will pay particular attention to composite penalties of
  the form $\phi(B x)$, where $B$ is a matrix corresponding to some constraint
  space, such as the discrete difference operator in fused Lasso.  
  
  The following concepts and definitions will be useful:

  \emph{Splitting} is a key tool that exploits an equivalence between the
  unconstrained optimisation problem and a constrained one that includes a
  latent--or slack--variable, $z$, where we write
  $$
  \min_x \left\{ l(x) + \phi(Ax) \right\}  
    \equiv  \min_{x,z} \left\{ l(x) + \phi(z) \right\} \; 
    \text{ subject to } \; z = A x \;.
  $$

  \emph{Envelopes} are another way of introducing latent variables. For example, we
  will assume that the objective $l(x)$ can take one of two forms of an envelope;  
  \begin{enumerate}
    \item a linear envelope
      $ l(x) = \sup_z \left\{x z - l^\star(z) \right\} $ 
      where $l^\star$ denotes the convex dual.
    \item
      a quadratic envelope
      $l(x) = \inf_z \left\{\half x^T \Lambda(z) x - \eta^T(z) x + \psi(z) \right\} $
     for some $ \Lambda, \eta, \psi $.
  \end{enumerate}

  The \emph{convex conjugate} of $l(x)$, $l^\star(z)$, is the point-wise supremum of a
  family of affine (and therefore convex) functions in $z$; it is convex even
  when $l(x)$ is not.  But if $l(x)$ is convex (and closed and proper),
  then the following dual relationship holds between $l$ and its conjugate:
  \begin{gather*}
    l(x) = \sup_{\lambda} \{ \lambda^T x - l^\star(\lambda) \} \text{ where }
    l^\star(\lambda) = \sup_{x} \{ \lambda^T x - l(x) \} \, .
  \end{gather*}
  If $l(x)$ is differentiable, the maximizing value of $\lambda$ is
  $\hat{\lambda}(x) = \nabla l(x)$.

  A function $g(x)$ is said to \emph{majorize} another function $f(x)$ at $x_0$ if
  $g(x_0) = f(x_0)$ and $g(x) \geq f(x)$ for all $x \neq x_0$.  If the same
  relation holds with the inequality sign flipped, $g(x)$ is said to be a
  \emph{minorizing} function for $f(x)$.  A $ \rho$-strong convex function satisfies
  $$
  f(x) \geq f(z) + u^\top (x-z) + \frac{\rho}{2} \| x-z \|^2_2, 
  \text{ where }  u \in \partial f(z)
  $$
  and $\partial$ denotes the \emph{subdifferential} operator defined by  
  $$
  \partial f(x) = \left\{v : f(z) \geq f(x) + v^T (z-x), 
    \forall z, x \in \operatorname{dom}(f)\right\} \;.
  $$
  A $ \rho $-smooth function satisfies
  $$
  f(x) \leq f(z) + \nabla f(z)^\top (x-z) + \frac{\rho}{2} \| x-z \|^2_2 , 
  \forall x,z \; .
  $$

  We also use the following conventions:
  $\sgn(x)$ is the algebraic sign of $x$, and $x_+ = \max(x, 0)$;
  $\iota_C(x)$ is the set indicator function taking the value $0$ if $x \in C$,
  and $\infty$ if $x \notin C$;
  $\Re^+ = [0, \infty)$, $\Re^{++} = (0, \infty)$, and $\overline{\Re}$ is the
  extended real line $\Re \cup \{-\infty, \infty\}$.

\section{Proximal operators and Moreau envelopes}
\label{sec:prox_operators}

  The key tools we employ are proximal operators and Moreau envelopes.
  Let $f(x)$ be a lower semi-continuous function, and let $\gamma > 0$ be a
  scalar.  The Moreau envelope $\moreau{\gamma}{f} (x)$ and proximal operator
  $\prox_{\gamma f} (x)$ with parameter $\gamma$ are defined as
  \begin{align}
    f^\gamma (x) &= \inf_{z } \left\{f(z) + \frac{1}{2\gamma} \enorm{z - x}^2  \right\}  \leq f(x) 
    \label{eq:moreau_envelope} \\
    \prox_{\gamma f} (x) &= \argmin_{z } \left\{  f(z)+ \frac{1}{2\gamma} \enorm{z - x}^2  \right\} \, .
    \nonumber
  \end{align}
  Intuitively, the Moreau envelope is a regularized version of $f$.  It
  approximates $f$ from below and has the same set of minimizing values
  \citep[Chapter 1G]{rockafellar:wets:1998}.  The proximal operator specifies
  the value that solves the minimization problem defined by the Moreau
  envelope.  It balances the two goals of minimizing $f$ and staying near $x$,
  with $\gamma$ controlling the trade-off.  
  Table~\ref{tab:prox}  provides an extensive list of closed-form solutions.

  \subsection{Properties of Proximal Operators}

  Our perspective throughout this paper will be to view proximal fixed point
  algorithm as the gradient of a suitably defined envelope function. By
  constructing different envelopes one can develop new optimisation algorithms.
  We build up to this perspective by first discussing the basic properties
  of the proximal operator and its relationship to the gradient of the standard
  Moreau envelope.  For further information, see \citet{parikh2013proximal} who
  provide interesting interpretations of the proximal operator.  Each one
  provides some intuition about why proximal operators might be useful in
  optimization.  We highlight three of these interpretations here that relate
  to the envelope perspective.
  
  First, the proximal operator behaves similarly to a gradient-descent step for
  the function $f$.  There are many ways of motivating this connection, but one
  simple way is to consider the Moreau envelope $\moreau{\gamma}{f}(x)$, which
  approximates $f$ from below.  Observe that the Moreau derivative is
  $$
  \partial \moreau{\gamma}{f}(x) = \partial \inf_{z } \left\{f(z) 
    + \frac{1}{2\gamma} \enorm{z - x}^2  \right\} = \frac{1}{\gamma}[x - \hat{z}(x)] \, 
  $$
  where $\hat{z}(x) = \prox_{\gamma f}(x)$ is the value that achieves the minimum.  
  Hence, 
  $$
  \prox_{\gamma f}(x) = x - \gamma \partial \moreau{\gamma}{f}(x) \, ,
  $$
  Thus, evaluating the proximal operator can be viewed as a gradient-descent step
  for a regularized version of the original function, with $\gamma$ as a
  step-size parameter.

  Second, the proximal operator generalizes the notion of the Euclidean
  projection.  To see this, consider the special case where $f(x) = \iota_C(x)$
  is the set indicator function of some convex set $C$.  Then 
  $\prox_f(x) = \argmin_{z \in C}  \| x-z \|_2^2$ is the ordinary Euclidean
  projection of $x$ onto $C$.  This suggests that, for other functions, the
  proximal operator can be thought of as a generalized projection.
  A constrained optimization problem $\min_{x \in C} f(x)$ has an equivalent
  solution as an unconstrained proximal operator problem. 
  Proximal approaches are, therefore, directly related to convex relaxation and quadratic
  majorization, through the addition of terms like $\frac{\rho}{2}\|x-v\|^2$ to
  an objective function--where $\rho$ might be a constant that bounds an operator or
  the Hessian of a function.  We can choose where these quadratic terms
  are introduced, which variables the terms can involve, and the order in
  which optimization steps are taken.  The envelope framework highlights such
  choices, leading to many distinct and familiar algorithms.  

  There is a close connection between proximal operators and
  fixed-point theory, in that $\prox_{\gamma f} (x^\star) = x^{\star}$ if and
  only if $x^{\star}$ is a minimizing value of $f(x)$.  To see this informally,
  consider the \textit{proximal minimization} algorithm, in which we start from
  some point $x_0$ and repeatedly apply the proximal operator:
  $$
  x^{t+1} = \prox_{\gamma f} (x^{t}) = x^t - \gamma \nabla \moreau{\gamma}{f}(x^t) \, .
  $$
  At convergence, we reach a minimum point $x^{\star}$ of the Moreau envelope,
  and thus a minimum of the original function.  At this minimizing value, we have
  $\nabla \moreau{\gamma}{f}(x^{\star}) = 0$ and thus $\prox_{\gamma f} (x^\star) =
  x^{\star}$.

  Finally, another key property of proximal operators is the Moreau decomposition for
  the proximal operator of $f^\star$, the dual of $f$: 
  \begin{align}
    x &= \prox_{\lambda f}(x) 
      + \lambda 
      \prox_{f^\star / \lambda}(\lambda x) 
      \nonumber \\
    I - \prox_{\lambda f}(x) &= \lambda \prox_{f^\star / \lambda}(\lambda x) 
      \label{eq:moreau_decomp}
  \end{align}
  The Moreau identity allows one to easily alter steps within a
  proximal algorithm so that some computations are performed in the dual (or primal) space. 
  Applications of this identity can also succinctly explain the relationship
  between a number of different optimization algorithms, as described in
  Section~\ref{sec:prox_algos_composite}.

  All three of these ideas---projecting points onto constraint regions, taking
  gradient-descent steps, and finding fixed points of suitably defined
  operators---arise routinely in many classical optimization algorithms.  It is
  therefore easy to imagine that the proximal operator, which relates to all
  these ideas, could also prove useful.

\subsection{Simple examples of proximal operators}

  Many intermediate steps in optimization problems can be written
  very compactly in terms of proximal operators of log likelihoods or penalty
  functions.  Here are two examples.

  Figure \ref{fig:moreau_envelope} provides a graphical depiction of these two
  concepts for the simple case $f(x) = |x|$.  In general the proximal operator
  may be set-valued, but it is scalar-valued in the special case where $f(x)$ is
  a proper convex function.

  \begin{Exa}
    Figure~\ref{fig:moreau_envelope} shows a simple proximal
    operator and Moreau envelope.  The solid black line shows the function 
    $f(x) = |x|$, and the dotted line shows the corresponding Moreau envelope
    $\moreau{}{f}(x)$ with parameter $\gamma=1$.  The grey line shows the function 
    $|x| + (1/2)(x-x_0)^2$ for $x_0 = 1.5$, whose minimum (shown as a red cross)
    defines the Moreau envelope and proximal operator.  This point has ordinate
    $\prox_f(x_0) = 0.5$ and abscissa $\moreau{}{f}(x_0) = 1$, and is closer than $x_0$
    to the overall minimum at $x=0$.  The blue circle shows the point 
    $(x_0, \moreau{}{f}(x_0))$, emphasizing the point-wise construction of the Moreau
    envelope in terms of a simple optimization problem.
    \begin{figure}[t]
    \begin{center}
    \includegraphics[width=3.25in]{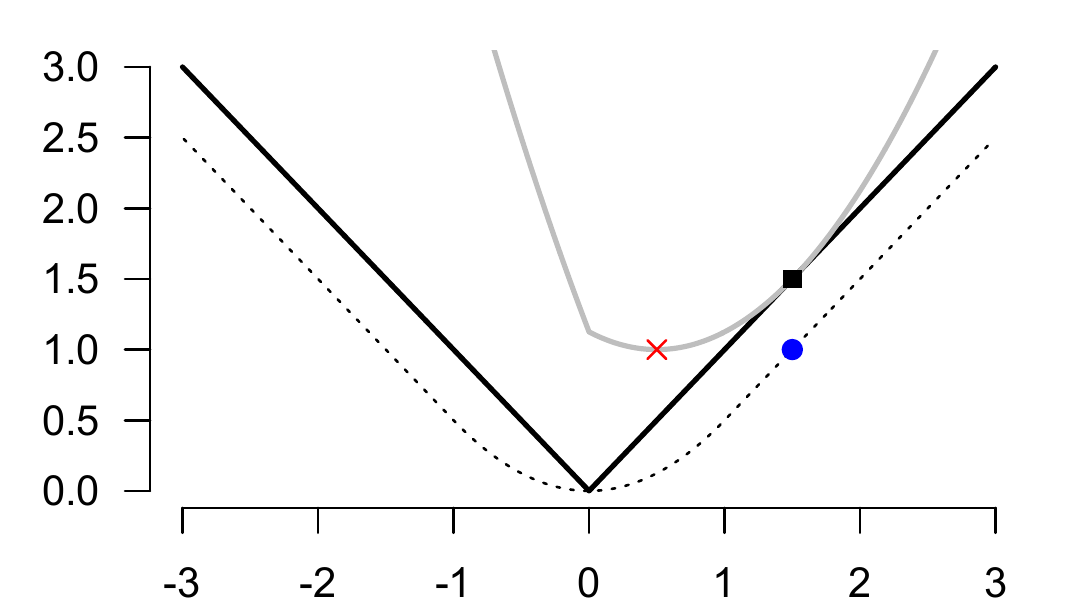}
    \caption{A simple example of the proximal operator and Moreau envelope.}
    \label{fig:moreau_envelope}
    \end{center}
    \end{figure}

    Let $\phi(x) = \lambda \| x \|_1$ and consider the proximal operator
    $\prox_{\gamma \phi}(x)$.  In this case the proximal operator is clearly
    separable in the components of $x$, and the problem that must be solved for
    each component is
    $$
    \min_{z \in \Re} \left\{ \lambda |z| + \frac{\gamma}{2} (z-x)^2 \right\} \, .
    $$
    This problem has solution
    \begin{equation}
    \label{eqn:softthresholding}
    \hat{z} = \prox_{\lambda|x|/\gamma} (x) = \sgn(x)(|x| - \lambda/\gamma)_+ = S_{\lambda/\gamma}(x) \, ,
    \end{equation}
    the soft-thresholding operator with parameter $\lambda/\gamma$.

  \end{Exa}

  \begin{Exa}
    Quadratic terms of the form
    \begin{equation}
    \label{eqn:quadratic_loglikelihood}
    l(x) = \frac{1}{2} x^T P x + q^T x + r \, ,
    \end{equation}
    are very common in statistics. They correspond to conditionally Gaussian
    sampling models and arise in weighted least squares problems, in ridge
    regression, and in EM algorithms based on scale-mixtures of normals.  For
    example, if we assume that $(y | x) \sim \N(Ax, \Omega^{-1})$, then 
    $l(x) = (y-Ax)^T \Omega (y-Ax) /2$, or
    $$
    P=A^T \Omega A \;, \quad q= - A^T \Omega y \;, \quad r=y^T \Omega y/2
    $$
    in the general form given above (\ref{eqn:quadratic_loglikelihood}).  If $l(x)$
    takes this form, its proximal operator (with parameter $1/\gamma)$ may be
    directly computed as
    $$
    \prox_{l/\gamma}(x) = (P+\gamma I)^{-1} (\gamma A^T  x - q) \, ,
    $$
    assuming the relevant inverse exists.  
  \end{Exa}

  General lesson: the proximal operator provides concise description of many
  iterative algorithms.  Practically useful only if the proximal operator can be
  evaluated in closed form or at modest computational cost.

\section{Proximal Algorithms}
  \subsection{The Proximal Gradient Method}
  \label{sec:basic_prox_algos}
 
  One of the simplest proximal
  algorithms is the proximal-gradient method which provides
  an important starting point for the more advanced techniques we describe
  in subsequent sections.  

  Suppose as in (\ref{eqn:canonicalform}) that the objective function is 
  $F(x) = l(x) + \phi(x)$, where $l(x)$ is differentiable but $\phi(x)$ is not.
  An archetypal case is that of a generalized linear model with a
  non-differentiable penalty designed to encourage sparsity.  The proximal
  gradient method is well suited for such problems.  It has only two basic
  steps which are iterated until convergence.
  \begin{description}
    \item[1) Gradient step.] 
      Define an intermediate point $v^t$ by taking a
      gradient step with respect to the differentiable term $l(x)$:
      $$
      v^t = x^t - \gamma \nabla l(x^{t}) \, .
      $$
    \item[2) Proximal operator step.] 
      Evaluate the proximal operator of the non-differentiable term $\phi(x)$
      at the intermediate point $v^t$:
      \begin{equation} 
        x^{t+1} = \prox_{\gamma \phi} (v^t) 
        = \prox_{\gamma \phi}\{ x^t - \gamma \nabla l(x^{t}) \} \, . 
        \label{eqn:prox_quad_alg}
      \end{equation} 
  \end{description}

  This can be motivated in at least two ways.

  \paragraph{As an MM algorithm.}

  Suppose that $l(x)$ has a Lipschitz-continuous gradient with modulus $\lambda_l$.
  This allows us to construct a majorizing function: whenever 
  $\gamma  \in (0, 1/\lambda_l]$, we have the majorization
  $$
  l(x) + \phi(x) \leq l(x_0) + (x - x_0)^T \nabla l(x_0) + \frac{1}{2\gamma} \enorm{x - x_0}^2 + \phi(x) \, ,
  $$
  with equality at $x=x_0$.  Simple algebra shows that the optimum value of the
  right-hand side is 
  $$
    \hat{x} = \argmin_x \left\{  \phi(x) + \frac{1}{2\gamma} \enorm{x - u }^2 \right\} \, , 
    \quad \text{ where } \quad u = x_0 - \gamma \nabla l(x_0) \, .
  $$
  This is nothing but the proximal operator of $\phi$, evaluated at an
  intermediate gradient-descent step for $l(x)$.

  The fact that we may write this method as an MM algorithm leads to the
  following basic convergence result.  Suppose that
  \begin{enumerate}
    \item $l(x)$ is convex with domain $\Re^n$.
    \item $\nabla l(x)$ is Lipschitz continuous with modulus $\lambda_l$, i.e.
    $$
    \enorm{\nabla l(x) - \nabla l(z)} \leq \lambda_l \enorm{x-z} \quad \forall x,z \, .
    $$
    \item $\phi$ is closed and convex, ensuring that $\prox_{\gamma \phi}$ makes sense.
    \item the optimal value is finite and obtained at $x^{\star}$.
  \end{enumerate}
  If these conditions are met, than the proximal gradient method converges at
  rate $1/t$ with fixed step size $\gamma = 1/\lambda_l$ \citep{beck2009gradient}.

  \paragraph{As the fixed point of a ``forward-backward'' operator.}  The
  proximal gradient method can also be interpreted as a means for finding the
  fixed point of a ``forward-backward'' operator derived from the standard
  optimality conditions from subdifferential calculus.  This has connections
  (not pursued here) with the forward-backward method for solving partial
  differentiable equations.  
  A necessary and sufficient condition that $x^{\star}$ minimizes $l(x)$ is that
  \begin{equation}
    \label{eqn:subdiffproxgrad}
    0 \in \partial \left\{ l(x^{\star}) + \phi(x^{\star})\right\} 
      = \nabla l(x^{\star}) + \partial \phi(x^{\star}) \, ,
  \end{equation}
  the sum of a point and a set.  We will use this fact to characterize
  $x^{\star}$ as the fixed point of the following operator:
  \begin{equation}
    \label{eqn:fixedpointproxgrad}
    x^{\star} = \prox_{\gamma \phi}\{ x^{\star} - \gamma \nabla l(x^{\star}) \} \, .
  \end{equation}
  To see this, let $I$ be the identity operator.  Observe that the optimality
  condition (\ref{eqn:subdiffproxgrad}) is equivalent to
  \begin{align*}
    0 &\in \gamma \nabla l(x^{\star}) - x^{\star} + x^{\star} + \gamma \partial \phi(x^{\star}) \\
    x^{\star} - \gamma \nabla l(x^{\star}) &\in x^{\star} + \gamma \partial \phi(x^{\star}) \\
    (I-\gamma \nabla l)x^{\star} &\in (I +  \gamma \partial \phi) x^{\star} \\
    x^{\star} &= (I +  \gamma \partial \phi)^{-1} (I-\gamma \nabla l) x^{\star} \\
              &=  \prox_{\gamma \phi} (x^{\star}-\gamma \nabla l(x^{\star})) \, ,
  \end{align*}
  the composition of two operators.  The final line appeals to the fact (see
  below) that the proximal operator is the resolvent of the subdifferential
  operator: $\prox_{\gamma \phi}(x) = (I + \gamma \partial \phi)^{-1} (x)$.
  Thus to find the solution, we repeatedly apply the operator having $x^\star$
  as a fixed point: 
  $$ 
  x^{t+1} = \prox_{\gamma^{t} \phi}\{ x^{t} - \gamma^{t}
  \nabla l(x^{t}) \} \, .  
  $$ 
  This is precisely the proximal gradient method.

  We now show that the proximal operator is the resolvent of the
  subdifferential operator.  By definition, if 
  $z \in (I + \gamma \partial l)^{-1} x$, then
  \begin{align*}
    x &\in (I + \gamma \partial l)z \\
    x &\in z + \gamma \partial l(z) \\
    0 &\in \frac{1}{\gamma}(z-x) + \partial l(x)  \\
    0 &\in \partial_z \left\{ \frac{1}{2\gamma}\enorm{z-x}^2 + l(x) \right\} \, .
  \end{align*}
  But for $0$ to be in the subdifferential (with respect to $z$) of the
  function on the right-hand side it is necessary and sufficient for
  $z$ to satisfy
  $$
  z = \argmin_u \left\{  \frac{1}{2\gamma}\enorm{u-x}^2 + l(u)  \right\} = \prox_{\gamma l}(x) \, .
  $$
  Therefore $z = \prox_{\gamma l}(x)$ if and only if $z \in (I + \gamma \partial l)^{-1} x$.  
  It is interesting that $(I + \gamma \partial l)^{-1}$
  is single-valued and therefore a function, even though $\partial l$ is
  set-valued.

  The proximal framework also applies to some non-convex regularisation penalties,
  e.g.  $L^q$-norm for $0 \leq q \leq 1$, for which we provide an example in
  Section~\ref{sec:L2_Lq_example}.

  \subsection{Iterative Shrinkage Thresholding}
  \label{sec:ist}

    Consider the proximal gradient method applied to a quadratic-form
    log-likelihood \eqref{eqn:quadratic_loglikelihood}, 
    as in a weighted least-squares problem, with a penalty function $\phi(x)$.  Then 
    $\nabla l(x) = A^T \Omega A x - A^T \Omega y$, and the proximal gradient
    method becomes 
    \begin{align*}
      x^{t+1} &= \prox_{\gamma^t \phi}\{ x^t - \gamma^{t} A^T \Omega ( A x^t - y )  \} \, .
    \end{align*}
    This algorithm has been widely studied under the name of IST, or iterative
    shrinkage thresholding \citep{fig:nowak:2003}.  
    Its primary computational costs at each iteration are: (1) multiplying the
    current iterate $x^t$ by $A$, and (2) multiplying the residual $A x^{t} - y$
    by $A^T \Omega$.  Typically the proximal operator for $\phi$ will be simple
    to compute, as in the case of a quadratic or $L^1$-norm/Lasso penalty, and will
    contribute a negligible amount to the overall complexity of the algorithm.

  \subsection{Proximal Newton}
  \label{sec:prox_newton}

    Proximal gradient, or forward-backward splitting, is a generalisation of the
    classical gradient approaches. They only require first-order
    information and their speed can be improved by using second order information,
    where the resulting algorithms mimic quasi-Newton procedures.  
    To do this, notice that the quadratic bound in \eqref{eqn:prox_quad_alg},
    implied by the definition of the proximal operator, implements a linear
    approximation of $l(x)$; however, one can, naturally, use higher order
    expansions to construct envelopes.
    If we let 
    \begin{align*}
      F_H(x,z) &= l(z) + {\nabla l(z)}^T (x-z) + \frac{1}{2} (x-z)^T H_z (x-z)
    \end{align*}
    Then we can calculate the proximal operators,
    \begin{align}
      \prox_{F_H}(z) &= z - \left(\gamma^{-1} I + H_z \right)^{-1} \nabla l(z)
      \label{eq:second_order_prox}
    \end{align}
    Instead of directly using the Hessian, $H_z = \nabla^2 l(z)$, approximations can be
    employed leading to quasi-Newton approaches.  The second-order bound, and
    approximations to the Hessian, are one way to interpret the half-quadratic
    (HQ) approach, as well as introduce quasi-Newton methods into the proximal
    framework.

    Proximal Newton methods are even possible for some non-convex problems; as in 
    \citep{chouzenoux2014variable} and Appendix~\ref{app:kl}. One advantage is
    that adding second-order derivative information can convexify some problems.


  \subsection{Nesterov Acceleration}

    One advantage of proximal algorithms is that we can accelerate the sequences
    within algorithms like \eqref{eqn:prox_quad_alg} by introducing an
    intermediate step that adds a momentum term to the slack variable, $z$,
    before evaluating the forward and backwards steps, 
    \begin{align*}
      z^{t+1} & = x^{t} + \theta_{t+1} ( \theta_{t}^{-1} -1 ) ( x^{t} - x^{t-1} )\\
      x^{t+1} & =  \prox_{\gamma^{-1} \phi}\left(z^{t+1} - \gamma^{-1} \nabla l(z^{t+1})\right) 
    \end{align*}
    with  $\theta_t = 2 / (t+1)$ and 
    $\theta_{t+1} ( \theta_{t}^{-1} -1 ) = (t-1)/(t+2)$.
    
    When $\phi$ is convex the proximal problem is strongly convex, and 
    advanced acceleration techniques can be used
    \citep{zhang2010regularized,meng2011accelerating}.

\section{Related Algorithms: ADMM, Divide and Concur, Bregman Divergences}
\label{sec:related_algos}

Many common estimation approaches can be interpreted as proximal point
or proximal gradient methods.  Much of the variation within these approaches
is simply due to the exact objective problem upon which a proximal algorithm
is being used.  In this section, we describe how splitting and functional
conjugacy results in new objective functions (or Lagrangians), which relate to
some well-known algorithms.  In Section~\ref{sec:prox_algos_composite} we
describe an overarching framework for the objective functions of these
algorithms and describe how proximal operators, their properties and resulting
algorithms are applied.

Our original problem $ \min_{x} l(x) + \phi(x) $ is clearly equivalent to
\begin{equation}
  \label{eqn:splitconstrained}
  \begin{gathered}
   \min_{x} \, l(z) + \phi(x) \\
   \text{subject to }\, x-z=0 \, ,
  \end{gathered}
\end{equation}
which we refer to as the ``primal'' problem.  We have introduced $z$ as a
redundant parameter (or ``slack variable''), and encoded a consensus
requirement in the form of the affine constraint $x-z=0$.

Other redundant parameterizations are certainly possible. For example, consider
the case of an exponential-family model for outcome $y$ with
cumulant-generating function $\psi(z)$ and with natural parameter $z$:
$$
p(y) = p_0(y) \exp \{ yz - \psi(z)\} \, .
$$
In a generalized linear model, the natural parameter for outcome $y_i$ is a
linear regression on covariates, $z_i = a_i^T x$. In this case $l(x)$ may be
written as
$$
l(x) = \sum_{i=1}^N l_i(x) \; \text{ where } \; l_i(x) =  \psi(a_i^T x) - y_i (a_i^T x)  \, ,
$$
up to an additive constant not depending on $x$.  Now introduce slack variables
$z_i = a_i^T x$. This leads to the equivalent primal problem
\begin{gather*}
 \min_{x, z} \, \sum_{i=1}^N \{\psi(z_i) - y_i z_i \} + \phi(x) \\
 \text{subject to }\, Ax - z = 0 \, .
\end{gather*}

These same optimization problems can arise when one considers 
scale-mixtures, or convex variational forms \cite{palmer2005variational}.  
The connection is made explicit by the dual function for a
density and its relationship with scale-mixture decompositions.  
For instance, one can obtain the following
equality for appropriate densities $p(x), q(z)$ and constants $\mu, \kappa$:
\begin{align*}
  -\log p(x) &= -\sup_{z>0} \log\left( p_N(x; \mu+\kappa/z, z^{-1}) q(z) \right)  \\
    &= \inf_{z > 0} \left\{ 
      \frac{z}{2}(x-\mu-\kappa/z)^2 - \log\left(\sqrt{z} q(z)\right) \right\}
      \; .
\end{align*}
where $p_N(x; \mu, \sigma^2)$ is the density function for a normal distribution
with mean $\mu$ and variance $\sigma^2$.
The form resulting from this normal scale-mixture envelope
is similar to the half-quadratic envelopes
described in Section~\ref{sec:envelopes}, and--more generally--the
objective in \eqref{eq:quad_objective}.
\citet{polson2014mixtures} describe these relationships in further detail.

The advantage of such a variable-splitting approach is that now the fit and
penalty terms are de-coupled in the objective function of the primal problem.
A standard tactic for exploiting this fact is to write down and solve the dual
problem corresponding to the original (primal) constrained problem.  This is
sometimes referred to as \emph{dualization}.  Many well-known references exist on this
topic \cite{bertsekas2011incremental}.  For this reason we focus on problem
formulation and algorithms for solving \eqref{eqn:splitconstrained}, avoiding
standard material on duality or optimality conditions.

The latent/slack variables allow us to view the problem of $ \min_x F(x)$ as
one of a joint minimisation of $ \min_{x,z} F(x,z) $ where the augmented
$F(\cdot, \cdot)$ can be easily minimisation in a conditional fashion.
Such alternating minimisation or iterated conditional mode (ICM)
\citep{besag1986statistical, csisz1984information} algorithms have a
long history in statistics.  The additional insight is that proximal
operators allow the researcher to perform the alternating minimisation step
for the non-smooth penalty, $\phi$, in an elegant closed-form fashion.
Moreover, Divide and Concur methods allow difficult high dimensional
problems to be broken down into a collection of smaller tractable subproblems
with the global solution being retrieved from the solutions to the
subproblems.

The following is a quick survey of some approaches that utilize variable
splitting and conjugacy.

\begin{description}

  \item[Dual Ascent]

    We first start with the simple problem
    $$ 
    \min_{x} \, l(x) \; 
      \text{subject to } \; A x = y 
    $$. We can solve this with a Lagrangian of the form 
    $$
    L(x, z) = l(x) + z^T(Ax - y) = l(x) + (A^T z)^T x - z^T y \, .
    $$
    The dual function is
    $g(z) = \inf_x L(x,z) = - l^{\star} (-A^T z) - y^T z $
    and the dual problem is
    $ \max_{z}\, g(z) $.

    Let $p^\star$ and $d^\star$ be the optimal values of the primal and dual
    problems, respectively.  Assuming that strong duality holds,
    the optimal values of the primal and dual problems are the same.
    Moreover, we may recover a primal-optimal point $x^\star$ from a dual-optimal
    point $z^\star$ via 
    $$
    x^\star = \argmin_x L(x, z^\star) 
      \quad \iff \quad 0 \in \partial_x L(x^\star, z^\star) \, .
    $$
    The idea of dual ascent is to solve the dual problem using gradient ascent via
    $$
    \nabla g(z) = \nabla_z  L(\hat{x}_z, z) \; , 
      \; \text{ where } \; \hat{x}_z = \argmin_x L(x, z) \, .
    $$
    The second term is simply the residual for the constraint: 
    $\nabla_z L(x, z) = Ax - y$.  Therefore, dual ascent involves
    iterating two steps: 
    \begin{align*}
      x^{k+1} &=  \argmin_x L(x, z^k) \\
      z^{k+1} &= z^k + \alpha_k (A x^{k+1} - y) 
    \end{align*}
    for appropriate step size $\alpha_k$.

  \item[Augmented Lagrangian]

    Take the same problem as before,
    $$ 
    \min_{x} \, l(x) \; \text{subject to } \; Ax = y
    $$
    with Lagrangian
    $L(x, z) = l(x) + z^T(Ax - y) $.

    The augmented-Lagrangian approach (or method of multipliers) seeks to stabilize
    the intermediate steps by adding a ridge-like term to the Lagrangian:
    $$
    L_{\gamma}(x, z) = l(x) + z^T(Ax - y) + \frac{\gamma}{2} \| Ax - y \|_2^2 \, .
    $$
    One way of viewing this is as the standard Lagrangian for the equivalent problem
    \begin{gather*}
      \min_{x} \, l(x) + \frac{\gamma}{2} \enorm{Ax-y}^2 \\
      \text{ subject to } Ax = y \, ,
    \end{gather*}
    For any primal-feasible $x$, the new objective remains unchanged, and thus has
    the same minimum as the original problem.
    The dual function is
    $g_{\gamma}(z) = \inf_x L_\gamma(x, z) $
    which is differentiable and strongly convex under mild conditions.  We can now
    use dual ascent for the modified problem, iterating
    \begin{align*}
      x^{k+1} &= \argmin_x  \left\{ l(x) + z^T(A x^k - y) 
        + \frac{\gamma}{2} \| Ax - y \|_2^2 \right\} 
        \\
      z^{k+1} &= z^k + \alpha_k (A x^{k+1} - y) \, .
    \end{align*}
    Thus the dual-variable update doesn't change compared to standard dual ascent.
    But the $x$ update has a regularization term added to it, whose magnitude
    depends upon the tuning parameter $\gamma$.  Notice that the step size $\gamma$ is used
    in the dual-update step.

    \paragraph{Scaled form.} Now re-scale the dual variable with 
    $u = \gamma^{-1} z$.  We can rewrite the augmented Lagrangian, with $r = Ax - y$, as
    \begin{align*}
      L_{\gamma}(x, u) &= l(x) + \gamma u^T (Ax - y) +  \frac{\gamma}{2} \| Ax - y \|_2^2 \\
      &= l(x) + \frac{\gamma}{2} \| r + u \|_2^2 - \frac{\gamma}{2} \| u \|_2^2
    \end{align*}
    This leads to the following dual-update formulas:
    \begin{align*}
      x^{k+1} &= \argmin_x \left\{ l(x) +  \frac{\gamma}{2} \| Ax - y + u^k \|_2^2 \right\} \\
      u^{k+1} &=  u^k + (A x^{k+1} - y) \, .
    \end{align*}
    Notice that the re-scaled dual variable is the running sum of the residuals
    $r^k = Ax^k - y$ from the primal constraint.  This is handy because the
    formulas are often shorter when working with re-scaled dual variables.

    \paragraph{Bregman iteration.}  The augmented Lagrangian method for solving
    $L^1$-norm/Lasso problems is called ``Bregman iteration'' in the compressed-sensing
    literature.  Here the goal is to solve the ``exact recovery'' problem via basis
    pursuit:
    \begin{gather*}
      \min_{x}\, \| x \|_1  \\ 
      \text{subject to } Ax = y \, ,
    \end{gather*}
    where $y$ is measured, $x$ is the unknown signal, and $A$ is a known ``short
    and fat'' matrix (meaning more coordinates of $x$ than there are observations).

    The scaled-form augmented Lagrangian corresponding to this problem is
    \begin{align*}
      L_{\gamma}(x, u) &= \| x \|_1 + \frac{\gamma}{2} \| Ax - y + u \|_2^2 
        - \frac{\gamma}{2} \|u \|_2^2 \, , 
    \end{align*}
    with steps
    \begin{align*}
      x^{k+1} &= \argmin_x  \left\{ \| x \|_1 +  \frac{\gamma}{2} \| Ax - z_k \|_2^2 \right\} \\
      z^{k+1} &=   y + z^k - A x^{k+1} \, ,
    \end{align*}
    where we have redefined $z^{k} = y - u^k$ compared to the usual form of the
    dual update.  Thus each intermediate step of Bregman iteration is like a lasso
    regression problem. (In the compressed sensing literature, this algorithm is
    motivated a different way, by appealing to Bregman divergences.  But it's the
    same algorithm.)

  \item[ADMM]

    Combining the ideas of variable splitting with the augmented
    Lagrangian one arrives at a method called ADMM (alternating-direction
    method of multipliers) for solving the problem \eqref{eqn:splitconstrained}.  The
    scaled-form augmented Lagrangian for this problem is
    $$
    L_{\gamma}(x, z, u) = l(z) + \phi(x) + \frac{\gamma}{2} \enorm{x - z + u}^2 
      + \frac{\gamma}{2} \enorm{u}^2  \, .
    $$
    ADMM is similar to Dual Ascent for this problem, except that we optimize
    the Lagrangian in $x$ and $z$ individually, rather than jointly, in each
    pass (hence ``alternating direction''):
    \begin{align*}
      z^{k+1} &= \argmin_z \left\{ l(z^{k}) +  \frac{\gamma}{2} \|x^k - z^{k} + u^k \|_2^2 \right\} \\
      x^{k+1} &= \argmin_x  \left\{ \phi(x^{k}) +  \frac{\gamma}{2} \| x^{k} - z^{k+1} + u^k \|_2^2 \right\}\\
      u^{k+1} &= u^k + x^{k+1} - z^{k+1}  \, .
    \end{align*}
    The first two steps are the proximal operators of $l(x)$ and
    $\phi(x)$, respectively.

    One way of interpreting ADMM is as a variant on the proximal gradient method
    for the dual problem corresponding to (\ref{eqn:splitconstrained}).  As a
    result, Nesterov-type acceleration methods may also be applied, with extra care
    to regularity conditions (in particular, strong convexity of $l(x)$).

\end{description}

\subsection{Bregman divergence and exponential families}

  Let $d(x)$ be a strictly convex differentiable function with convex/Legendre dual $b$.
  The Bregman divergence from $x$ to $y$ induced by $d$ is
  $$
  D_{d}(x, y) = d(x) - d(y) - d'(y) (x-y) \geq 0 \, .
  $$
  This is the vertical distance between $d(y)$ and the extrapolated ``guess'' for
  $d(y)$ based on the tangent line at $x$.  In the multivariate case everything
  carries through with gradients/planes replacing derivatives/lines.

  There is a unique Bregman divergence associated with every exponential family.
  It corresponds precisely to the relationship between the natural
  parameterization and the mean-value parameterization.  Suppose that
  $$
  p(y; \theta) = p_0(y) \exp \{ y \theta - b(\theta) \} \, .
  $$
  The expected value of $y$, as a function of $\theta$, is given in terms of the
  cumulant-generating function as $\mu(\theta) = b'(\theta)$.  This is sometimes
  referred to as Tweedie's formula \citep{robbins1964empirical, efron2011tweedie}, 
  and has come up repeatedly in a variety of different contexts.
  By the envelope formula, the maximizing value of $\mu$ in the second equation
  (as a function of $\theta)$ satisfies $\mu(\theta) = b'(\theta)$.  This lets us
  recognize the dual variable $\mu$ as the mean-value parameterization; that is,
  the natural and mean-value parameterizations form a Legendre pair.

  Hence, we can write an exponential-family model as either: (1) in
  terms of the natural parameter $\theta$ and the cumulant-generating function
  $b$, $p(y; \theta, b)$; or (2) in terms of the mean-value parameter $\mu$ and
  the Bregman divergence induced by the Legendre dual $d = b^{\star}$, $p(y ;
  \mu, d)$.

  \paragraph{Splitting on the mean-value parameter.}  
  Another use of variable splitting is when we write the model in terms of its mean-value
  parameterization:
  $$
  p(y_i ; \mu_i(x) ) \propto \exp\{-D_{d}[y_i,  \mu_i(x)]\} \, ,
  $$
  where $d = b^{\star}$ and $\mu_i(x) = E(y_i ; x)$ is the expected value, given the parameter.

  Assuming we are still using the canonical link, we may now write the model in
  terms of a penalized Bregman divergence with split variables: 
  \begin{gather*}
    \min_{x, z}\, \sum_{i=1}^N D_{d}(y_i, z_i) + \phi(x) \\
    \text{subject to } \mu(a_i^T x) - z_i = 0 \, .
  \end{gather*}
  where $\phi(x)$ is the penalty function.

  \begin{Exa}[Poisson regression]

    In a Poisson model $y_i \sim \operatorname{Pois}(\mu_i)$, $\mu_i = \exp(\theta_i)$ for
    natural parameter $\theta_i = a_i^T x$.  The cumulant generating function is
    $b(\theta) = \exp(\theta)$, and thus $d(\mu) = \mu \log \mu - \mu$.  After
    simplification, the divergence $D_d(y, \mu) = \mu - y \log \mu + (\mu - y)$.  The
    optimization problem can then be split as
    \begin{gather*}
      \min_{x, z}\, \sum_{i=1}^N (z_i - y_i \log z_i) + \phi(x) \\
      \text{subject to } a_i^T x  = \log z_i \, .
    \end{gather*}

  \end{Exa}

\subsection{Divide and Concur}
\label{sec:divide_concur}
  Divide and Concur provides a general approach to hierarchical
  statistical models that require optimisation of a sum of $J$ composite
  functions of the form
  $$
  \max_{x \in \mathcal{X}} \sum_{j=1}^{J+1} l_j(A_j x ) + \phi(B x)
  $$
  DC adds slack variables, $z_j$ for $j \in [1,\dots,J+1]$, to
  ``divide'' the problem together with equality constraints so that the solutions ``concur''.
  We have the equivalent constrained optimization problem
  $$
  \max_{x,z} \sum_{j=1}^{J+1} l_j(z_j)
  \text{ under constraints $z_j=A_j x$, $z_{J+1}= B x$.} 
  $$
  where $l_{J+1} = \phi$, $A_{J+1} = B$.
  This can be solved using an iterative proximal splitting algorithm (e.g. multiple ADMM,
  split Bregman).  Specifically, under ADMM \citep{parikh2013proximal} one finds, with $\bar{x}^t = \frac{1}{J+1} \sum_{j=1}^{J+1} x^t_j$, that
  \begin{align*}
    x^{t+1}_j &= \prox_{\lambda l_j \circ A_j}(\bar{x}^t - u_j^k) \\
    u^{t+1}_j &= u_j^{t} + x^{t+1}_j - \bar{x}^{t+1} \;.
  \end{align*}
  Divide and Concur 
  \citep{gravel2008divide} methods are a natural approach to big data problems as they
  break a hard high-dimensional problem into tractable, independently
  computable sub-problems via splitting and then find the global solution from
  the solutions to each sub-problem.
 
\section{Envelope Methods}
\label{sec:envelopes}

  In this section we introduce different types of envelopes: the
  \emph{forward-backward} envelope (FBE), \emph{Douglas-Rachford} envelope
  (DRE), and the \emph{half-quadratic} (HQ) envelope, and \emph{Bregman
  divergence} envelopes.  Within this framework, new algorithms are generated
  as a gradient step of an envelope
  Section~\ref{sec:prox_algos_composite}
  dissects these envelopes, shows their relationship to Lagrangian approaches,
  and provides a framework within which they can be derived and extended.
  
  \subsection{Forward-Backward Envelope}
  
  Suppose that we have to minimise $ F = l + \phi $ where $l$ is
  strongly convex and possesses a continuous gradient with
  Lipschitz constant $\lambda_l$ so that $ | \nabla^2 l(x) | \leq \lambda_l $.
  The penalty $\phi$ is only assumed to be proper lower semi-continuous and
  convex.  If we don't have an ``exact'' quadratic envelope (see the discussion
  in \ref{sec:quad_composite}), then we can argue as follows.  
  
  First, we define the FBE, $F^{\text{FB}}_\gamma (x)$, which will possess some desirable
  properties (see \citet{patrinos2013proximal}). 
  \begin{align*}
    F^{\text{FB}}_\gamma ( x) & \defeq \min_v \left \{ l(x) + {\nabla l(x)}^T (v-x) + \phi ( v ) 
      + \frac{1}{2 \gamma} \vnorm{v-x}^2 \right \} \\
    & = l(x) - \frac{\gamma}{2} \vnorm{\nabla l(x)}^2 
      + \moreau{\gamma}{\phi} \left ( x - \gamma \nabla l(x) \right ) 
  \end{align*}
  
  If we pick $ \gamma \in ( 0 , \lambda_l^{-1} ) $, the matrix $ I - \gamma
  \nabla^2 l(x)$ is symmetric and positive definite. The
  stationary points of the envelope $ F^{\text{FB}}_\gamma (x)$ are the solutions
  $x^\star$ of the original problem which satisfy
  $ x = \prox_{\gamma \phi}( x - \gamma \nabla l(x) )$.
  This follows from the derivative information 
  $$
  \nabla F^{\text{FB}}_\gamma (x) = ( I - \gamma \nabla^2 l(x) ) G_\gamma (x) 
    \text{ where } G_\gamma (x) = \gamma^{-1} ( x - P_\gamma ( x )  ) 
  $$
  where $ P_\gamma (x) = \prox_{\gamma \phi}( x - \gamma \nabla l(x) ) $.
 
  With these definitions, we can establish the descent property for the FBE
  \begin{align*}
    F^{\text{FB}}_\gamma (x ) & \leq F(x) - \frac{\gamma}{2} \vnorm{G_\gamma (x)}^2\\
    F(P_\gamma (x)) & \leq F^{\text{FB}}_\gamma (x) 
      - \frac{\gamma}{2} (1 - \gamma \lambda_l ) \vnorm{G_\gamma(x)}^2 \;.
  \end{align*}
  Hence for $ \gamma \in ( 0 , \lambda_l^{-1} ) $ the envelope value always decreases
  on application of the proximal operator of $\gamma \phi$ and we can determine
  the stationary points.  See Appendix~\ref{app:convergence} for further details.
  
 
 
  \subsection{Douglas-Rachford Envelope}
 
  Mimicking the forward-backward approach,
  \citet{patrinos2014douglas} derive the Douglas-Rachford envelope (DRE)
  \begin{align*}
    F_\gamma^{\text{DR}} (x) & = \moreau{\gamma}{l}(x) - 
    \frac{\gamma}{2} \| \nabla \moreau{\gamma}{l}(x) \|_2^2 + 
      \phi^\gamma \left( x - 2 \gamma \nabla \moreau{\gamma}{l}(x) \right)
      \\
      & = \min_z \left\{ l(x^\star) 
      + \nabla l(x^\star)^\top (z- x^\star) + \phi(z) + 
      \frac{1}{2 \gamma} \| z- x^\star \|^2 \right\} \; .  
  \end{align*}
  where $\moreau{\gamma}{l}$ is, again, the Moreau envelope of the function $l$
  and $x^\star = \prox_{\gamma l}(x)$.
  This can be interpreted as a backward-backward envelope and is a special case
  of a FBE evaluated at the proximal operator of $\gamma l$, namely
  $$
  F_\gamma^{\text{DR}} (x) = F_\gamma^{FB} \left ( \prox_{\gamma l}(x) \right  ) \; .
  $$
  Again the gradient of this envelope produces the following proximal algorithm
  (see \citet{patrinos2014douglas}) 
  which converges to the solution to $ \min_x \left\{ l(x) + \phi(x) \right\} $
  given by the iterations 
  \begin{align*}
  w^{t+1} & = \prox_{\gamma l}(x^t)\\
  z^{t+1} & = \prox_{\gamma \phi}(2 w^t - x^t)\\
  x^{t+1} & = x^t + ( z^t-w^t )
  \end{align*}
  There are many ways to re-arrange the DR algorithm. For example, with an
  intermediate variable, $v=w-x$, we could equally well iterate
  \begin{equation*}
  w^{t+1} = \prox_{\gamma l} (x^t-v^t) \; , \;   x^{t+1} 
    = \prox_{\gamma \phi} (w^t + v^t) \; , \; v^{t+1} 
    = v^t + ( w^t-x^t ) \; .
  \end{equation*}
 
  \subsection{Half-Quadratic Envelopes}
 
  We now provide an illustration of a quasi-Newton algorithm within the class
  of Half-Quadratic (HQ) optimization problems 
  \citep{geman1995nonlinear,geman1992constrained}.  
  This envelope applies to the commonly used $L^2$-norm where $l(x) = \| A x - y\|^2$,
  and can be used in conjunction with some non-convex $\phi$.
  See \citet{nikolova2005analysis} for convergence rates and comparisons of the
  different algorithms.
  
  The half-quadratic envelope (HQE) is defined by
  \begin{align*}
    F^{\text{HQ}}(x) & = \inf_v \left \{ Q(x,v) + \psi(v) \right \} \\
    \text{ where } Q(x,v) & = v x^2 \text{ or } (v-x)^2
  \end{align*}
  and the function, $Q(x,v)$, is half-quadratic in the variable $v$. 
  In the HQ framework, the term $\psi(v)$ is usually understood to be 
  the convex conjugate of some function, e.g. $\psi(v) = \phi^\star(x)$. 
  \begin{Exa}

    Suppose that we wish to minimise the functional
    $$
    F(x) = \frac{1}{2} \vnorm{A x - y}^2 + \gamma \Phi(x)  
      \text{ where }  \Phi(x) = \sum_{i=1}^d \phi( ( B^T x - b )_i ) 
    $$
    and we're given $\phi(x) = F^{\text{HQ}}(x)$.
    Then we need to solve the joint criterion
    $$
    F(x, v)  =  \frac{1}{2} \|A x - y|^2 
      + \gamma \sum_{i=1}^d Q( \delta_i , v_i ) 
      + \gamma \sum_{i=1}^d \psi( v_i) \; . 
    $$                    
    where $\delta_i = ( B^T x - b )_i $.
    There is an equivalence between gradient linearisation and quasi-Newton.
    These algorithms give the iterative mappings: 
    $$ 
    x^{t+1} = L( \hat{v} (x^{t} ) )^{-1} A^T y \text{ and } 
    x^{t+1} = x^{t} -  L(x^{t})^{-1} \nabla_x F( x^{t} ) , 
    $$
    where $L(x^{t})$ is a step size function.  
    They are identical, with derivative information 
    \begin{align*}
      \nabla_x F(x) & = A^T A x - A^T y + \gamma \sum_{i=1}^d B_i 
        \frac{ \phi^\prime ( \vnorm{\delta_i} ) }{ \vnorm{\delta_i} } B_i^T x 
        \\
        & =  ( A^T A + \gamma B \operatorname{V}(x) B^T ) x - A^T y 
        \\
        & = L(\hat{v}(x)) x - A^T y 
    \end{align*}
    for $\operatorname{V}(x)  = diag ( \hat{v} (  \vnorm{\delta}_{i=1}^d ) )$ and
    $L(\hat{v}(x)) = A^T A + v B\, \operatorname{V}(x) B^T$.

    Here  $ \hat{v} (x) = \phi^\prime (x) / 2 x $ for Geman-Yang (GY) and
    $\hat{v} ( x ) = x - \phi^\prime (x) $ for 
    Geman-Reynolds (GR).


  \end{Exa}

  \subsection{Bregman Divergence Envelopes}

  Many statistical models, such as those generated by an exponential family
  distribution, can be written in terms of a Bregman divergence.  One is then
  faced with the joint minimisation of an objective function of the form 
  $ D(x,v) + \phi(x) + \psi(v) $. To minimise over $(x,v)$ we can use an alternating
  Bregman projection method. To perform the minimisation of $v$ given $x$ we
  can make use of the $D$-Moreau envelope which is defined by
  \begin{equation}
    \moreau{D}{\phi}(x) = \inf_v \left\{ D(x,v) + \phi(v) \right\}
    \nonumber
  \end{equation}
  where $ D(x,v)$ is a Bregman divergence, $D(x,v) \geq 0 $ and attains
  equality at $x=v$.  The Bregman divergence has a three-point law of cosines
  triangle inequality, which helps to establish descent in proximal algorithms
  (see Appendix~\ref{app:convergence}).  Many commonly used EM and MM
  algorithms in statistics and variational Bayes models use envelopes of this
  type.
  
  The key insight is that the proximal operator generated by the $D$-Moreau
  envelope allows one to add non-smooth regularisation penalties to
  traditional exponential family models. In our applications, we illustrate
  this with logistic and Poisson regression both of which can be interpreted as
  Bregman divergence measures of fit in the objective function.

  We now turn to the general case of a quadratic envelope with a composite
  regularization penalty.
  
   
\section{Proximal Algorithms for Composite Functions}
\label{sec:prox_algos_composite}

  Building off the general objective in \eqref{eq:general_objective},
  we now consider the composite objective given by the optimisation
  $$
  \min_x F(x) \defeq l(x) + \phi(B x) \;.
  $$
  Composite mappings of the form, $\phi(B x)$, arises in multi-dimensional
  statistical models that account for structural constraints or correlations,
  making such terms both common and important consideration in the construction
  and estimation of a model.  Therefore, any practical framework for estimating
  statistical models must be capable of addressing these mappings somewhat broadly.   
  The methodology described here uses splitting, proximal operators and Moreau
  envelopes.  We find that this combination of tools can be used together 
  easily, applies to a broad range of functions and underlies many state-of-the-art
  approaches that scale well in high dimension.

  We start by noting that many optimization approaches, including the ones in
  Section~\ref{sec:related_algos}, can be summarized by listing the
  general forms of the objective functions/Lagrangians that result from
  splitting and duality:
  \begin{align*}
    \text{primal} && F(x) &= l(x) + \phi(B x) \\
    \text{primal-dual} && F_{PD}(x,z) &= l(x) + z^T(B x) - \phi^\star(z)  \\ 
    \text{split primal} && F_{SP}(x,w,z) &= l(x) + \phi(w) + z^T(B x - w) \\
    \text{split dual} && F_{SD}(x,w,z) &= l^\star(w) + \phi^\star(z) + x^T(-B^T z - w) 
  \end{align*}
  The motivation for using the primal-dual and the split forms (see
  \citet{esser2010general}) lies in how they decouple $\phi$ from $B$ without
  affecting its solution to the primal problem $\min_x F(x)$.  We refer to these
  re-formulations of the primal objective function, and their implied 
  minimization/maximization requirements, as joint \emph{objective problems}.  
  The exact objective problems given above are by no means exhaustive and
  need not apply to only one function in the primal objective.
  
  As mentioned in Section~\ref{sec:related_algos},
  the split problems can be viewed as Lagrangian formulations that each arise
  separately from the definition of the convex conjugate or Fenchel dual, and
  relate to each other, in the general case, by the Max-Min inequality
  \citep{boyd2009convex}
  $$
  \sup_q \inf_v F(q,v) \leq \inf_v \sup_q F(q,v)
  $$
  In the special case of closed proper convex functions, we have the following 
  $$
    \min_x F(x) = \min_x \sup_z F_{PD}(x,z) 
    = \max_z \min_{x,w} F_{SP}(x,w,z) 
    = \max_x \min_{z,w} F_{SD}(x,w,z) \;,
  $$
  made possible for the dual problems by noting the equality when $\phi$ is convex 
  $$
  \phi(B x) = \sup_z \left\{ z^T B x - \phi^\star(z) \right\}  \;.
  $$
  In this case, $F_{SP}(x,w,z)$ and $F_{PD}(x,z)$ are equated by
  \begin{align*}
    \min_{w \geq 0} F_{SP}(x,w,z) &= \min_{w \geq 0}\left\{ \phi(w) + l(x) + z^T(B x - w) \right\} \\
                                  &= l(x) + z^T B x + \min_{w \geq 0}\left\{ \phi(w) - z^T w \right\} \\
                                  &= l(x) + z^T B x - \phi^\star(z) \\
                                  &= F_{PD}(x,z)
  \end{align*}
  The solutions $x^\star$,$w^\star$, and $z^\star$ can also be the results of
  proximal operators.

  \subsection{Proximal Solutions within Objective Problems}

  Given an objective problem, one must specify the
  exact steps to solve the sub-problems within it, i.e. the problems in $w$
  and/or $z$.  In some cases, closed forms solutions for the primal or dual
  functions (i.e. $l(x), l^\star(w), \phi(x), \phi^\star(z)$) in some variables might
  not be available, or computationally efficient; however,
  exact solutions to related problems that share the same critical points may
  be easily accessible.  These related problems, or the entire objective
  problem, can take the form of the envelopes
  in Section~\ref{sec:envelopes} and, as a result, the solutions for
  terms in the objective can be proximal operators.  In fact, the envelope
  representation can be seen as a way to represent--altogether--the combination
  of an objective problem and the solutions to each of its
  latent/slack/splitting terms as proximal operators.

  Especially in cases where multiple majorization steps are taken (to solve for--say--$w$
  and $z$ in a $F_{SP}$ problem) the use of proximal operators,
  their properties, and the associated fixed-point theory can simplify otherwise
  lengthy constructions and convergence arguments.  As well, using 
  the proximal operator's properties, like the Moreau identity, one can move
  easily between the different objective problems and, thus, primal and dual
  spaces.  It is also worth mentioning that the efficacy of certain
  acceleration techniques can depend on the objective problem (see
  \citet{beck2014fast}) and, similarly, the proximal steps taken. 
  
  For a further connection to the general optimization literature,
  the quadratic term in the proximal operator can be seen as a
  quadratic penalty for a linear constraint in a Lagrangian.  When a split
  objective is used, the application of a proximal operator results in an
  objective function that is very similar--or equivalent--to an augmented Lagrangian.
  Specifically, the addition of a squared
  term in the $F_{SP}$ problem leads to the ADMM estimation technique
  in which one iterates through conditional solutions to $x$ and $z$ at each step,
  with solutions given by proximal points. 
  Both \citet{parikh2013proximal} and \citet{chen1994proximal} observe that, for the
  splitting/composite problem, the augmented Lagrangian for ADMM is
  \begin{equation}
    \begin{gathered}
      \phi(w) + l(x) + z^T(B x-w) + \frac{\rho}{2} \|Ax - z\|^2 \\
      = F_{SP}(x,w,z) + \frac{\rho}{2} \|A x - z\|^2 
    \end{gathered}
    \label{eq:aug_lagr_admm}
  \end{equation}
  We can consider direct proximal solutions to some variables in this objective;
  for instance, $z^\star = \prox_{F_{SP}(x,w,z)/\rho}(A x)$.  The objective is then
  \begin{gather*}
    F_{SP}(x,w,z^\star) + \frac{\rho}{2} \|A x - z^\star\|^2 
  \end{gather*}
  If $z^\star$ as a function of $w$ is linear, then it may be possible to take another
  proximal step, like $w^\star = \prox_{F_{SP}(x,w,z^\star)/\rho}(\dots)$.

  We aren't restricted to using the proximal operators directly implied
  by an objective problem, such as those that appear when $l, l^\star$ and/or 
  $\phi, \phi^\star$
  are--or contain--quadratic terms in their arguments. 
  Instead, one can apply a surrogate or approximation
  (e.g. envelopes, majorization/minorization) to terms within an
  objective problem and effectively induce a proximal operator.  
  This could be done for the purposes of imposing or approximating
  a constraint, as in \eqref{eq:aug_lagr_admm}, or even for approximating
  solutions to such a constraint.
  Notice that the proximal step producing $z^\star$ for \eqref{eq:aug_lagr_admm}
  involves a composite argument, $A x$.  When exact solutions to the composite
  proximal operator aren't available, one can consider ``linearizing''
  $\frac{\rho}{2} \|A x - z\|^2$ with $\frac{\rho}{2 \lambda_A} \|x - z\|^2$,
  where
  $ \sigma_{\text{max}}(A^T A) \leq \lambda_A $, yielding
  \begin{gather*}
    F_{SP}(x,w,z) + \frac{\rho}{2} \|A x - z\|^2 \leq
    F_{SP}(x,w,z) + \frac{\rho}{2 \lambda_A} \|x - z\|^2 \;. 
  \end{gather*}
  This approach can be seen as a simple majorization, and, when combined
  with the proximal solution for $z$, as a forward-backward envelope
  for the sub-problem.  Implementations of this approach include the linearized
  ADMM technique, or the split inexact Uzawa method, and are described in the
  context of Lagrangians by \citet{chen1994proximal} and primal-dual algorithms in 
  \citet{chambolle2011first}.  \citet{magnusson2014convergence} details
  splitting methods in terms of augmented-Lagrangians for non-convex
  objectives.  

  To demonstrate the framework described here, we give an example of how
  proximal operators, their properties, and these concepts can be used to
  derive an algorithm for a specific objective problem. 
  \begin{Exa}
    For proper, convex $l(x), \phi(x)$ with Lipschitz continuous derivatives, we 
    start with the primal-dual problem 
    $$
      \max_z \inf_x \left\{ l(x) + z^T (Bx) - \phi^\star(z) \right\}
    $$
    and notice that the $\argmin$ for the sub-problem in $x$, $l(x)+ z^T (B x)$, is
    given by the fixed point, for $\lambda_l > 0$, 
    $$
    x^\star = \prox_{\lambda_l (l(x)+ z^T B x)}(x^\star) \;.
    $$ 
    By a property of proximal operators, namely 
    \begin{equation}
      \prox_{g(z) + u^T z}(q) = \prox_{g}(q-u) \; ,
      \label{eq:prox_translate}
    \end{equation}
    for a generic function $g(z)$ and variables $q$, $z$ and $u$
    (obtained by completing the square in the definition of the operator) we have 
    $$
    x^\star = \prox_{\lambda_l (l(x) + z^T B x)}(x^\star) = \prox_{\lambda_l l}(x^\star - \lambda_l B^T z) \;.
    $$
    Now, we're left with only the sub-problem in $z$,
    \begin{gather*}
      \max_z \left\{ l(x^\star) + z^T (B x^\star) - \phi^\star(z) \right\} 
      = -\min_z \left\{ \phi^\star(z) - z^T (B x^\star) - l(x^\star) \right\} 
      \;.
    \end{gather*}
    We can take yet another proximal step, for the minimization
    problem, $\phi^\star(z) - z^T (B x^\star)$, in $z$ with constant $\lambda_\phi$. 
    Using \eqref{eq:prox_translate} and \eqref{eq:moreau_decomp}, we find that the
    $\argmin$ satisfies
    \begin{align*}
      z^\star &= \prox_{\lambda_\phi \phi^\star}(z^\star + \lambda_\phi B x^\star) 
    \end{align*}

    Next, let's say we find that the proximal solution to $\phi^\star$ is
    problematic in some cases, yet the solution $x^\star$ is still desirable.
    Using the Moreau decomposition in \eqref{eq:moreau_decomp}, we can easily
    derive an alternative for those cases:
    \begin{align*}
      \prox_{\lambda_\phi \phi^\star}(z^\star + \lambda_\phi B x^\star)
      &= \frac{1}{\lambda_\phi} \left(I - \prox_{\phi/\lambda_\phi}\right) 
        \circ \left( \lambda_\phi (z^\star +  B x^\star) \right)
    \end{align*}
    Hence, we have the following implied iterative algorithm:
    \begin{equation}
      \begin{aligned}
        x^\star &= \prox_{\lambda_l l}(x^\star- \lambda_l B^T z^\star) \\
        z^\star &= \frac{1}{\lambda_\phi} \left(I - \prox_{\phi/\lambda_\phi}\right)
          \circ \left( \lambda_\phi (z^\star +  B x^\star) \right)
      \end{aligned}
      \label{eq:prox_comp_algo}
    \end{equation}

  \end{Exa}

  If we further separate the last step in \eqref{eq:prox_comp_algo} into two steps
  and simplify by setting $\lambda_l = \lambda_\phi = 1$, we arrive at 
  \begin{align*}
    x^\star &= \prox_{l}(x^\star - B^T u^\star) \\ 
    w^\star &= \prox_{\phi}(u^\star + B x^\star) \\
    u^\star &= u^\star - (w^\star - B x^\star) \;.
  \end{align*}
  This has the basic form of techniques like alternating split Bregman, ADMM,
  split inexact Uzawa, etc., which demonstrates how versatile the proximal operator
  and it's properties are when applied to the broad class of objective problems. 
  The differences between approaches often involve assumptions on $l$ and $\phi$,
  such as Lipschitz continuity, and the exact order of steps.  See
  \citet{chen2013primal} for more details.

\subsection{General Quadratic Composition}
\label{sec:quad_composite}

  Consider, now, the most general form of a quadratic objective
  \begin{equation}
    \argmin_x \inf_{z} \left\{ F_\Lambda (x,z) = 
    \frac{1}{2} x^T \Lambda(z) x - \eta^T(z)x  + \phi(B x) \right\}
      \label{eq:quad_objective}
  \end{equation}
  where $\Lambda(z) > 0$.
  Again, such forms can arise when one majorizes with a second-order approximation of
  $l(x)$ around $z$.  This also makes \eqref{eq:quad_objective} the Moreau
  envelope defined in \eqref{eq:moreau_envelope}.  The general quadratic case, in which
  $\Lambda(z)$ is not necessarily diagonal, can be addressed with splitting
  techniques.  
  

  This form, when $\Lambda(z)$ is symmetric positive definite, encompasses the
  approaches of \citet{geman1995nonlinear,geman1992constrained}
  Assuming $B$ is positive definite, a proximal point solution can be obtained
  by setting $l(x) = x^T \Lambda(z) x - \eta^T x$ in
  \eqref{eq:prox_comp_algo}.  The general solution to a quadratic-form proximal
  operator--like \eqref{eqn:quadratic_loglikelihood}--is, again, given by
  \begin{align*}
    \prox_{\lambda_l l(x)}(q) &= \left(I + \lambda_l \Lambda(z)\right)^{-1}(q+\lambda_l \eta)
  \end{align*}
  which, together with the split-dual formulation, implies a proximal point
  algorithm of the form 
  \begin{align*}
    x^\star &= \prox_{\lambda_l l(x)}(x^\star - \lambda_l B^T z^\star) \\
    &= \left(I + \lambda_l \Lambda(z^\star)\right)^{-1}(x^\star - \lambda_l B^T z^\star + \lambda_l \eta) \\
    z^\star &= \frac{1}{\lambda_\phi} \left(I - \prox_{\phi/\lambda_\phi}\right)
      \circ \left( \lambda_\phi (z^\star +  B x^\star) \right)
  \end{align*}

  We've now introduced the sub-problem of solving the following
  system of linear equations:
  $$
  \left(I + \lambda_l \Lambda(z)\right) q^\star = (q+\lambda_l \eta) \;.
  $$ 
  Using the exact solution to the system of equations would
  reflect methods that involve Levenberg-Marquardt
  steps, quasi-Newton methods, and Tikhonov regularization, and is related
  to the use of second-order Taylor approximations to an objective function.
  Naturally, the efficiency of computing exact solutions depends very much on the 
  properties of $I+\lambda_l \Lambda(z)$, since the system defined by this
  term will need to be solved on each iteration of a
  fixed point algorithm.  When $\Lambda(z)$ is constant, a decomposition can be
  performed at the start and reused, so that solutions are computed quickly at
  each step.  For some matrices, this can mean only $O(n)$ operations per
  iteration.  In general, however, the post-startup iteration cost is $O(n^2)$.

  Other approaches, like those in \citet{chen2013primal, argyriou2011efficient} 
  do not attempt to directly solve the aforementioned system of equations.  Instead
  they use a forward-backward algorithm on the dual objective, $F_{PD}$.
  For simplicity, let $\Lambda(z) = A$ be symmetric positive definite,
  and $A = R^T R$ its Cholesky decomposition.  The Cholesky decomposition
  won't be a required component in the resulting implied algorithm; it is
  used here for a simplified exposition.

  Starting with the split-dual objective for $l(x) = \half x^T A x - \eta^T x$,
  \begin{align*}
    \min_x \max_z &\left\{\half x^T A x - \eta^T x + z^T B x - \phi^\star(z) \right\} \\
    = \min_x \max_z &\left\{ \frac{1}{2} 
      \|R x - R^{-1}(\eta - B^T z)\|^2 
      - \frac{1}{2} \|R^{-1}(\eta - B^T z)\|^2 - \phi^\star(z) \right\} \;.
  \end{align*}
  A solution for the problem in $x$ is easily obtained from the proximal operator 
  and is $x^\star = A^{-1}(\eta - B^T z)$, or from the second line we could still
  arrive at the same solution via a first-order--or linearized-- quadratic
  bound inspired by the 2-norm inequality $\| M v\| \leq \|M\| \|v\|$.  
  That is  
  \begin{align*}
    \|R x - R^{-1}(\eta - B^T z)\|^2 &\leq \|R\|^2 \|x - A^{-1}(\eta - B^T z)\|^2 \\
                                     &\leq \sigma_{\text{max}}(A) \|x - A^{-1}(\eta - B^T z)\|^2
  \end{align*}
  Now, at $x = x^\star$ we have the following problem in $z$:
  \begin{gather*}
    \max_z \left\{ -\frac{1}{2} \|R^{-1}(\eta - B^T z)\|^2 - \phi^\star(z) \right\}
    = \min_z \left\{  
      \frac{1}{2} \|R^{-1} B^T z - R^{-1} \eta \|^2 + \phi^\star(z) \right\} 
  \end{gather*}
  Again, we can use a forward-backward proximal solution to the above problem, where
  $l(z) = \frac{1}{2} \|R^{-1} B^T z - R^{-1} \eta \|^2$, so that
  $$
  \nabla l(z) =  B R^{-T}\left(R^{-1} B^T z - R^{-1} \eta \right) 
  = \lambda_2 \left( B A^{-1} B^T z - B A^{-1} \eta \right) \;,
  $$ 
  Then, with $\lambda_2 \geq \sigma_{\text{max}}(B A^{-1} B^T)/2$, we can obtain
  $z^\star$ as the proximal solution
  \begin{align}
    z^\star &= \prox_{\lambda_2 \phi^\star}(z - \lambda_2 \nabla l(z)) 
    \nonumber \\
    z^\star &= \prox_{\lambda_2 \phi^\star}(z - \lambda_2 \left(B A^{-1} B^T z + B A^{-1} \eta \right)) 
    \nonumber \\
     &= \left(I - \prox_{\lambda_2^{-1} \phi}\right) \circ
    \left(\left(I - \lambda_2 B A^{-1} B^T \right) z + B A^{-1} \eta\right) 
    \label{eq:fb_comp_prox_alg_z}
  \end{align}
  In sum, we have an implied proximal point algorithm similar to
  \eqref{eq:prox_comp_algo} that is, instead, based on a first-order
  forward-backward method.

  \begin{Exa}
    A related example of this variety of split forward-backward algorithm is
    used by \citet{argyriou2011efficient}, who apply Picard-Opial iterations
    given by
    $$
    H_k = \kappa I + (1- \kappa) H \;, 
    $$ 
    for $\kappa \in (0,1)$, to find a fixed point, $v^\star$, of the operator
    $$
      H(v) \defeq \left ( I - \prox_{\gamma^{-1} \phi} \right ) 
        \left ( B A^{-1} \eta + ( I - \gamma B A^{-1} B^T ) v \right ) 
        \; \;  , \forall v \in \Re^p
    $$
    where $ 0 < \gamma < 2 / \sigma_{max} \left ( B A^{-1} B^T \right ) $.
    The operator $H$ is understood to be non-expansive, so, by Opial's theorem,
    one is guaranteed convergence, and, when $H$ is a contraction, this
    convergence is linear.  After finding $v^\star$, one sets
    $x^\star = A^{-1} \left ( \eta - x B^T v^\star \right ) $.
    
    Noting the similarities with \eqref{eq:fb_comp_prox_alg_z}, we see that
    $v$ here can be interpreted as the dual variable $z$.  What distinguishes
    this approach from others is that there are fewer upfront restrictions on 
    the matrix operator $B$.  
    \citet{chen2013primal} discuss the number of iterations, $k$, in the
    process of finding the fixed point $v^\star$ and detail a one-step
    algorithm with similar scope.   

  \end{Exa}

\section{Applications}
\label{sec:applications}

\subsection{Logit loss plus Lasso penalty}
\label{sec:logit_l1_example}

  To illustrate our approach, we simulate observations from the model
  \begin{align*}
    (y_i|p_i) &\sim \operatorname{Binom}(J, p_i) \\
    p_i &= \operatorname{logit}^{-1}(a_i^T x)
  \end{align*}
  where $i = 1,\dots,100$, $a_i^T$ is a row vector of $A \in \mathbb{R}^{100 \times 300}$,
  $x \in \mathbb{R}^{300}$ and $J=2$.  The $A$ matrix is simulated from $\operatorname{N}(0,1)$ 
  variates and normalized column-wise.  The signal $x$ is also simulated from 
  $\operatorname{N}(0,1)$ variates, but with only $10\%$ of entries being non-zero.

  Here $m_i$ are the number of trials, $y_i$ the number of successes and
  $m=\sum_{i=1}^n m_i$ the total number of trials in the classification problem.
  The composite objective function for sparse logistic regression is then given by 
  \begin{align*}
    \argmin_x \sum_{i=1}^n \left\{ m_i \log(1+e^{a_i^Tx}) - y_i a_i^T x \right\} 
    + \lambda \sum_{j=1}^p |x_j|
  \end{align*}
  To specify a proximal gradient algorithm all we need is an envelope such
  as those commonly used in Variational Bayes.  In this example, we use the simple
  quadratic majorizer with Lipschitz constant $\lambda$ given by 
  $\|A^T A\|_2/4 = \sigma_{\text{max}}(A)/4$, and a penalty coefficient
  $\lambda$ set to $0.1 \sigma_{\text{max}}(A)$.
   
\begin{figure}[t]
    \includegraphics[scale=0.50]{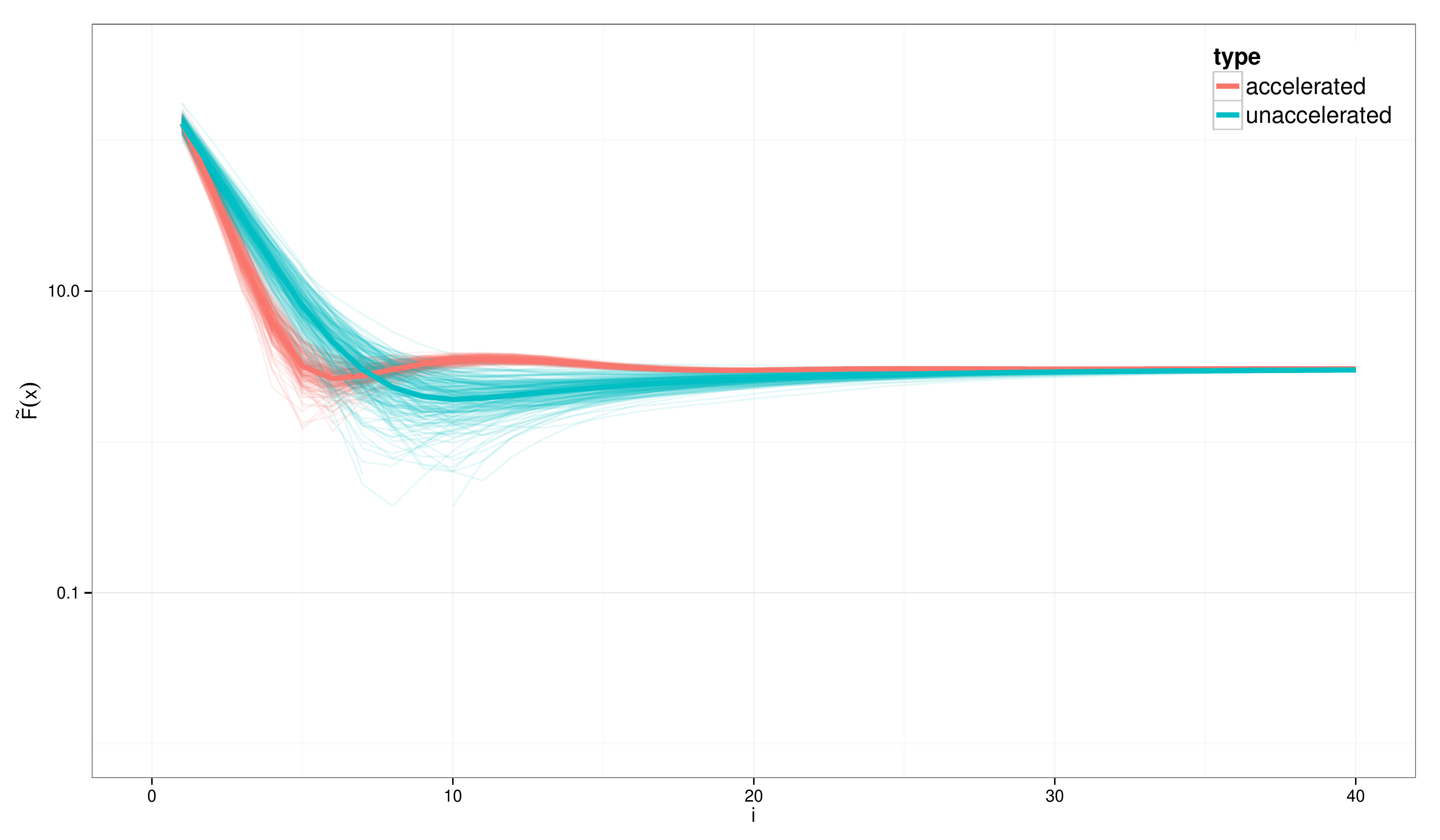}
    \caption{(Adjusted) objective values for iterations of the proximal gradient
      method, with and without acceleration, applied to a logistic regression
      problem with an $L^1$-norm penalty.}
    \label{fig:logit_l1_obj}
  \end{figure}

  Figure~\ref{fig:logit_l1_obj} shows the (adjusted) objective values
  per iteration with and without Nesterov acceleration.  We can see the non-descent nature of the algorithm and the clear advantage
  of adding acceleration.

\subsection{Logit Fused Lasso}
\label{sec:logit_l1_comp_example}

  To illustrate a logit fused lasso problem, we compare a Geman-Reynolds
  inspired quadratic envelope
  for the multinomial logit loss and a fused lasso penalty with the standard
  Lipschitz-bounded gradient step.  We define the following quantities
  \begin{align*}
    \Lambda(v) &= 2 \sum_{i=1}^n m_i \lambda(a_i^T v) a_i a_i^T 
               = 2 A^T \diag({\bf m} \cdot \lambda(A v)) A \\
    \eta^T &= 2 \sum_{i=1}^n \left(y_i - m_i/2\right)a_i^T \;.
  \end{align*}
  Now we compute $x_t$, conditional on $w$, for the envelope
  \begin{align*}
    \sum_{i=1}^n \left\{ m_i \log(1+e^{a_i^Tx}) - y_i a_i^T x \right\} 
    + \| D^{(1)} x \|_1 &=
    \min_y \left\{ \half x^T \Lambda(w) x - \eta^T x + c(w) 
    + \gamma \|D^{(1)} x \|_1 \right\} 
  \end{align*}
  To do this, we employ the Picard-Opial composite method of \citet{argyriou2011efficient}. 

\begin{figure}[t]
    \includegraphics[scale=0.50]{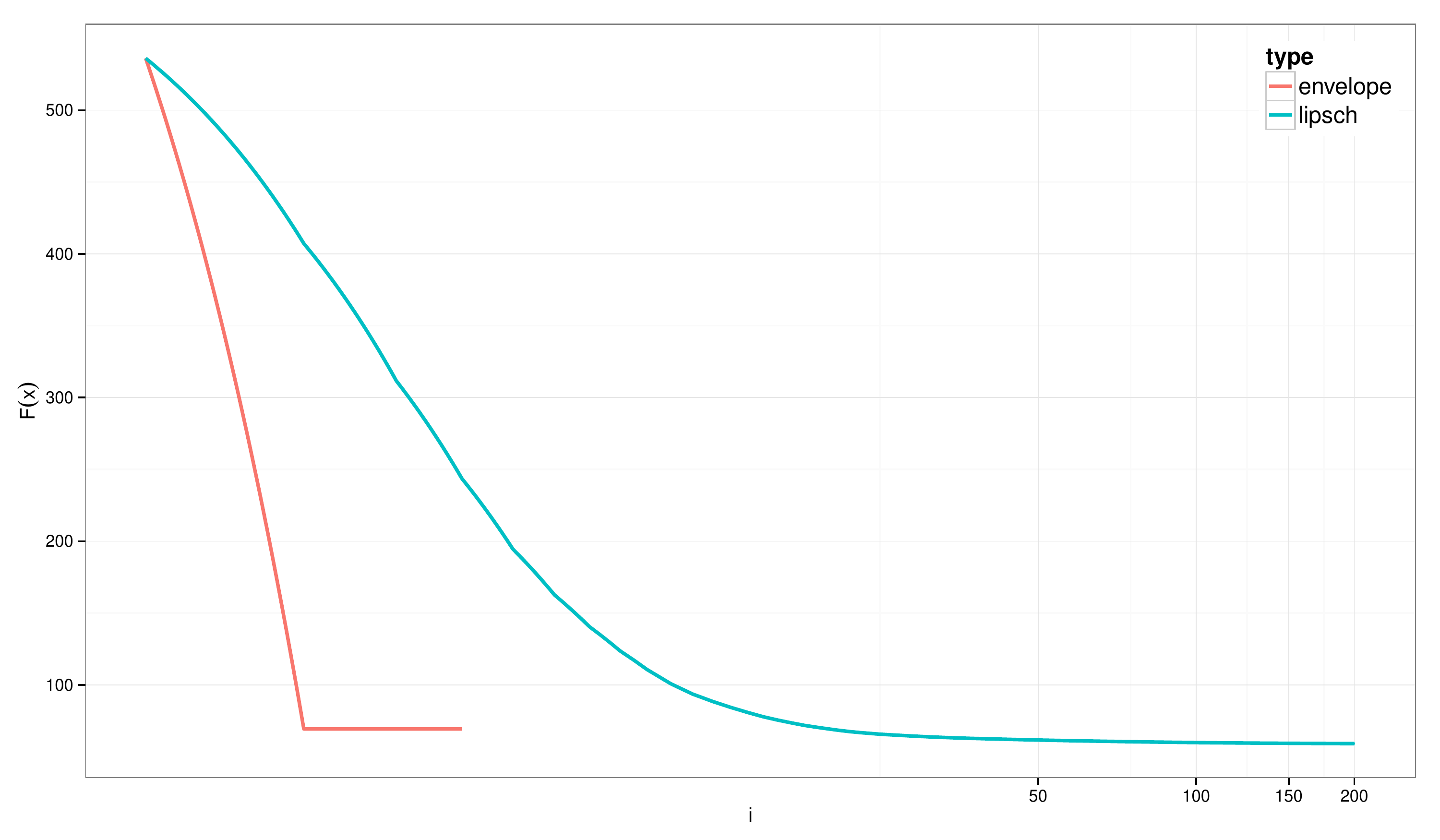}
    \caption{Objective values for iterations of two proximal composite
      formulations applied to a multinomial logistic regression
      problem with a composite $L^1$-norm penalty.  Both are run until 
      the same numeric precision is reached. }
    \label{fig:logit_comp_l1_obj}
  \end{figure}

  Simulations were performed in a similar fashion as Section~\ref{sec:logit_l1_comp_example}
  but with $N=100$, $M=400$, $m=2$ and where $D^{(1)} x$ has a fused lasso
  construction consisting of first-order differences of $x$.  
  Figure~\ref{fig:logit_comp_l1_obj} show the objective values
  for iterations of each formulation.  With the use of second-order
  information, we have extremely fast convergence to the solution.

  For data pre-conditioning, we perform the following decompositions:
  $A = U \Sigma V^T$, the singular value decomposition (SVD),
  $\Lambda^{-1}(v) = \half A^{-1} D^{-1} A^{-T}$, where
  $D = \diag({\bf m} \cdot \lambda(A v)) $.  This implies that
  one SVD of $A$, or generalized inverse, is required to compute
  all future $\Lambda^{-1}(v)$ and thus providing computational savings.

\subsection{Poisson Fused Lasso}
\label{sec:poisson_fused_example}

  To illustrate an objective that is not Lipschitz, but still convex, we use
  a Poisson regression example with a fused lasso penalty.  We simulated a
  signal given from the model
  \begin{align*}
    (y | x) &\sim \operatorname{Pois}(\exp(A x)) \\
    \phi(x) &= \|D^{(1)} x \|_1 = \sum_{j=1}^p |x_j - x_{j-1}| 
  \end{align*}
  In our simulation, the true sparse parameter vector $x$ has
  $10$\% non-zero signals from $\operatorname{N}(0,1)$.
  The design matrix $A \in \Re^{100 \times 300}$ is also generated from
  $\operatorname{N}(0,1)$, then column normalized. 

  In sum, we have a negative log-likelihood and regularization penalty
  of the composite form
  \begin{equation*}
    F(x) = \sum_{i=1}^n \exp(a_i^T x) - y_i a_i^T x 
    + \sum_{j=1}^p |x_j - x_{j-1}| = \sum_{i=1}^n \exp(a_i^T x) 
    - y_i a_i^T x + \| D^{(1)} x \|_1 \; .
  \end{equation*}
  where $a_i$ are the column vectors of $A$ and $D^{(1)} x$ is the matrix
  operator of first-order differences in $x$. 
  Since the Poisson loss function is not Lipschitz, but still convex, we replace
  the constant gradient step with a back-tracking line search.  This can be
  accomplished with a back-tracking line search step.

\begin{figure}[t]
    \includegraphics[scale=0.50]{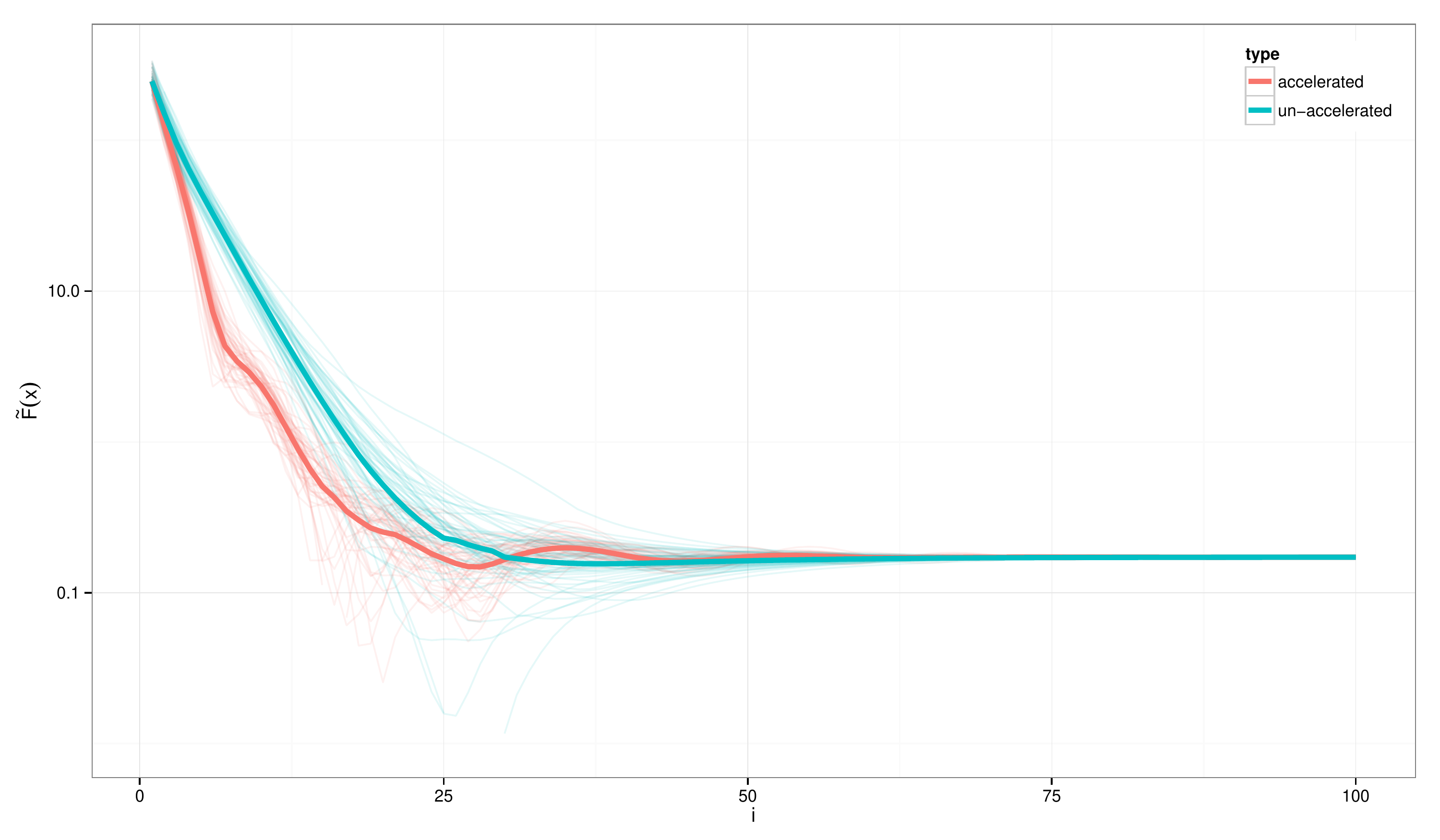}
    \caption{(Adjusted) objective values for iterations of the proximal gradient
      method, with and without acceleration, applied to a Poisson regression
      problem with a fused $L^1$-norm penalty.}
    \label{fig:pois_fused_l1_obj}
  \end{figure}

  Figure~\ref{fig:pois_fused_l1_obj} shows the objective value results for
  each method, with and without acceleration.
  An alternative approach is given by \citet{green1990use}, who describes an
  implementation of an EM algorithm for penalised likelihood estimation.

  \subsection{\texorpdfstring{$L^2$}{L2}-norm loss plus \texorpdfstring{$L^q$}{Lq}-norm penalty for \texorpdfstring{$0<q<1$}{0<q<1}}
  \label{sec:L2_Lq_example}

  A common non-convex penalty is the bridge norm, $L^q$-norm for $0 < q < 1$.
  There are a number of ways of developing a proximal algorithm to solve
  such problems.  The proximal operator of $L^q$-norm has a closed-form, multi-valued
  solution and convergence results are available for proximal methods
  in \citet{marjanovic2013exact} and \citet{attouch2013convergence}.  For this example,
  we choose the former approach.

  The regularization problem involves find the minimizer of an $L^2$-norm loss with an
  $L^q$-norm penalty for $0 < q < 1$,
  $$
  \hat{x}^q_\lambda \defeq 
  \argmin_x \left\{ \frac{1}{2} \vnorm{y-A x}^2 + \lambda \sum_{j=1}^p |x_i |^q \right\} \; ,
  $$
  The component-wise, set-valued proximal $L^q$-norm operator is given by
  \begin{align*}
    \prox_{\lambda \phi_q}(y) &=
      \begin{cases}
        0 & \text{ if } |y| < h_\lambda \\ 
        \{0, \sgn(y) x_\lambda \} & \text{ if } |y| = h_\lambda \\ 
        \sgn(y) \hat{x} & \text{ if } |y| > h_\lambda \\ 
      \end{cases}
  \end{align*}
  where 
  \begin{align*}
    b_{\lambda,q} &= \left(2 \lambda (1-q)\right)^{\frac{1}{2-q}} \\
    h_{\lambda,q} &= b_{\lambda,q} + \lambda q b_{\lambda,q}^{q-1} \\
    \hat{x} + \lambda q \hat{x}^{q-1} &= |y|, \hat{x} \in (b_{\lambda, q},|x|) 
  \end{align*}
  \citet{attouch2013convergence}
  describe how the objective for this problem is a Kurdyka-\L ojasiewicz (KL) 
  function, which provides convergence results for an inexact (multi-valued proximal operator) 
  forward-backward algorithm given by
  $$
  x^{t+1} \in \prox_{\lambda \gamma_t \|\cdot\|_p}
    \left( x^t - \gamma_t (A^T A x^t - A^T b) \right) \;.
  $$
  Interestingly, the KL convergence results for forward-backward splitting on
  appropriate non-convex continuous functions bounded below imply that the
  solution choice for multi-valued proximal maps--as in the $L^q$-norm case--does
  not affect the convergence properties.  See Appendix~\ref{app:kl} for more
  information.

  An alternative approach is the variational representation of the $L^q$-norm;
  however, this doesn't satisfy the convergence conditions of 
  \citet{allain2006global} within the half-quadratic framework. 

  \citet{marjanovic2013exact} detail how 
  cyclic descent can be used to apply the proximal operator
  in a per-coordinate fashion under a squared-error loss.  The
  cyclic descent method is derived from the following algebra.  First,
  a single solution to the squared-error loss minimization problem can be given
  for a component $i$ of $x$, by
  \begin{equation*}
    0 = \nabla_i l(x) = A_i^T ( A x - y) = A_i^T  ( A_i  x_i + A_{-i} x_{-i} - y)
  \end{equation*}
  where $A_i$ is column $i$ of $A$, and $A_{-i}, x_{-i}$ have column/element $i$
  removed.  Applied to a quadratic majorisation scheme we find that at iteration $t$
  \begin{equation*}
    x^{t+1}_i = \frac{A_i^T(y-A_{-i} x_i^{t+1} )}{A_i^T A_i} 
      = \frac{A_i^T r^{t}}{\|A_i\|^2} + x^{t}_i
  \end{equation*}
  with $y - A x^{t} = r^{t}$.
  In a similar fashion to gradient descent, this
  involves $O(n)$ operations for updates of $A_i^T r^{t}$, so one cycle
  is $O(n p)$.

  We simulate a data vector $ y \in \Re^n $ from a regression model 
  $$
  y = A x + \sigma \epsilon \; \text{ where } \; \epsilon \sim \operatorname{N}(0,1)
  $$
  with an underlying sparse parameter value $ x \in \Re^d $ with $n=100, d=256$, 
  in which the true sparse $x$ has $5$\% non-zero signals generated
  from $\operatorname{N}(0,1)$. The design matrix $A \in \Re^{100 \times 256}$ 
  is also generated from $\operatorname{N}(0,1)$
  then column normalized. We set the signal-to-noise ratio at $16.5$ to match the
  simulated example from \citet{marjanovic2013exact} which gives $\sigma=0.0369$.

\begin{figure}[t]
    \includegraphics[scale=0.50]{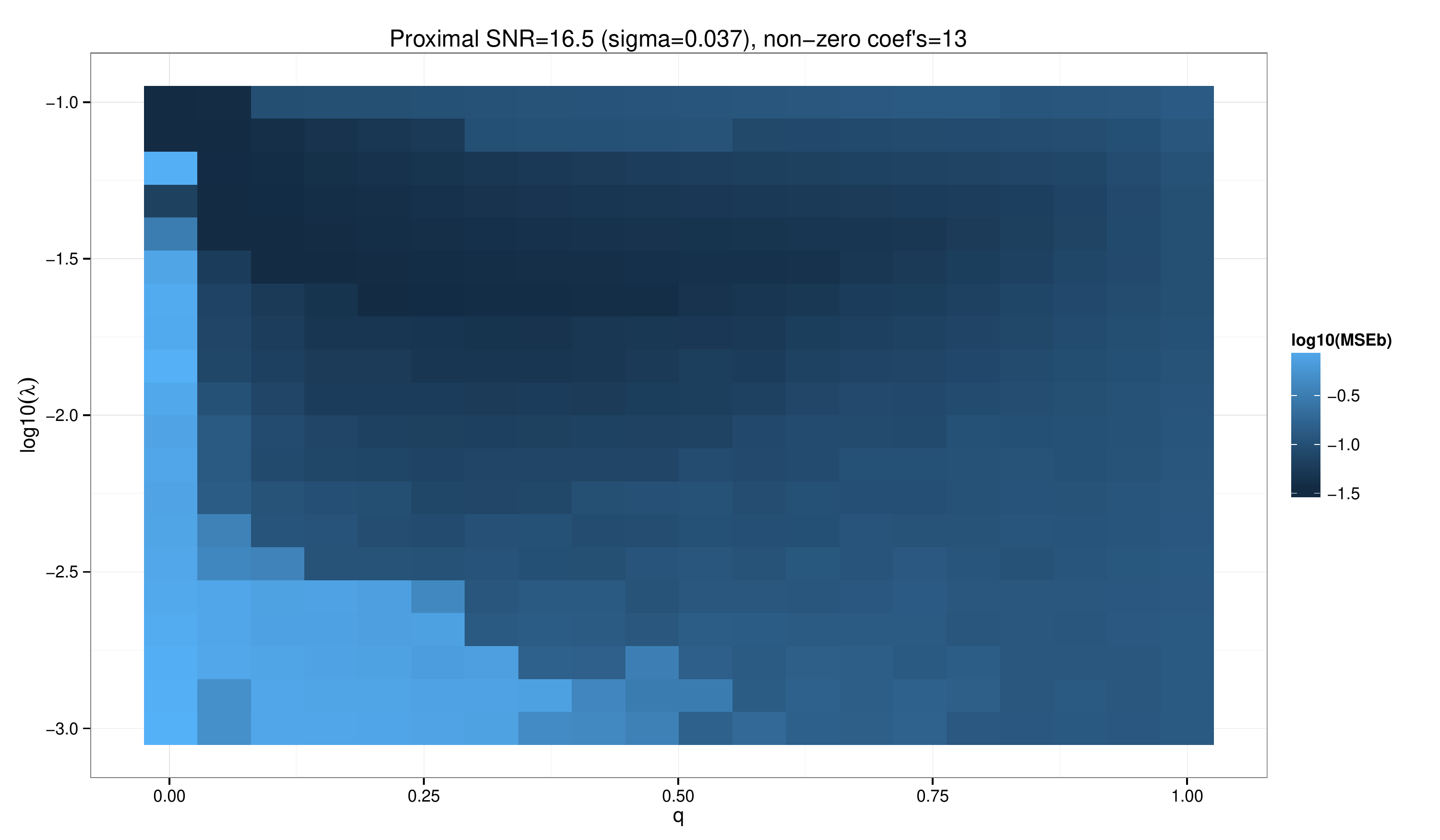}
    \caption{Penalty weight, $\lambda$, vs. MSE and $q$ for a 
      $L^2$-norm error with an $L^q$-norm penalty, $0 <q<1$, estimated
      via cyclic descent and proximal solutions.}
    \label{fig:ms_sim}
  \end{figure}

  Figure~\ref{fig:ms_sim} plots the mean squared error (MSE) versus
  the log-regularisation penalty and the power in the $L^q$-norm penalty.
  Essentially, this consists of contours of $ \log_{10} ( \text{MSE} ( \hat{x} ) ) $
  on a plot of $0<q<1$ versus the amount of regularization $\log_{10} (\lambda )$.
  One interesting feature of this model is that 
  the estimated regression coefficients $\hat{x}^q_\lambda$ can 
  jump to sparsity as $0 < q < 1$, and this will be illustrated in a regularized path
  for the next example.

\subsection{Prostate Data}
\label{sec:prostate_example}

  As a practical example  of our methodology, we consider the 
  prostate cancer dataset, which examines the relationship
  between the level of a prostate specific antigen and a number of clinical 
  factors.
  The variables are log cancer volume ({\tt lcavol}), log prostate
  weight ({\tt lweight}), age ({\tt age}), log of the amount of benign prostatic hyperplasia
  ({\tt lbph}), seminal vesicle invasion ({\tt svi}), log of capsular penetration ({\tt lcp}),
  Gleason score ({\tt gleason}), and percent of Gleason scores 4 or 5 ({\tt pgg45}).

\begin{figure}[t]
    \includegraphics[scale=1]{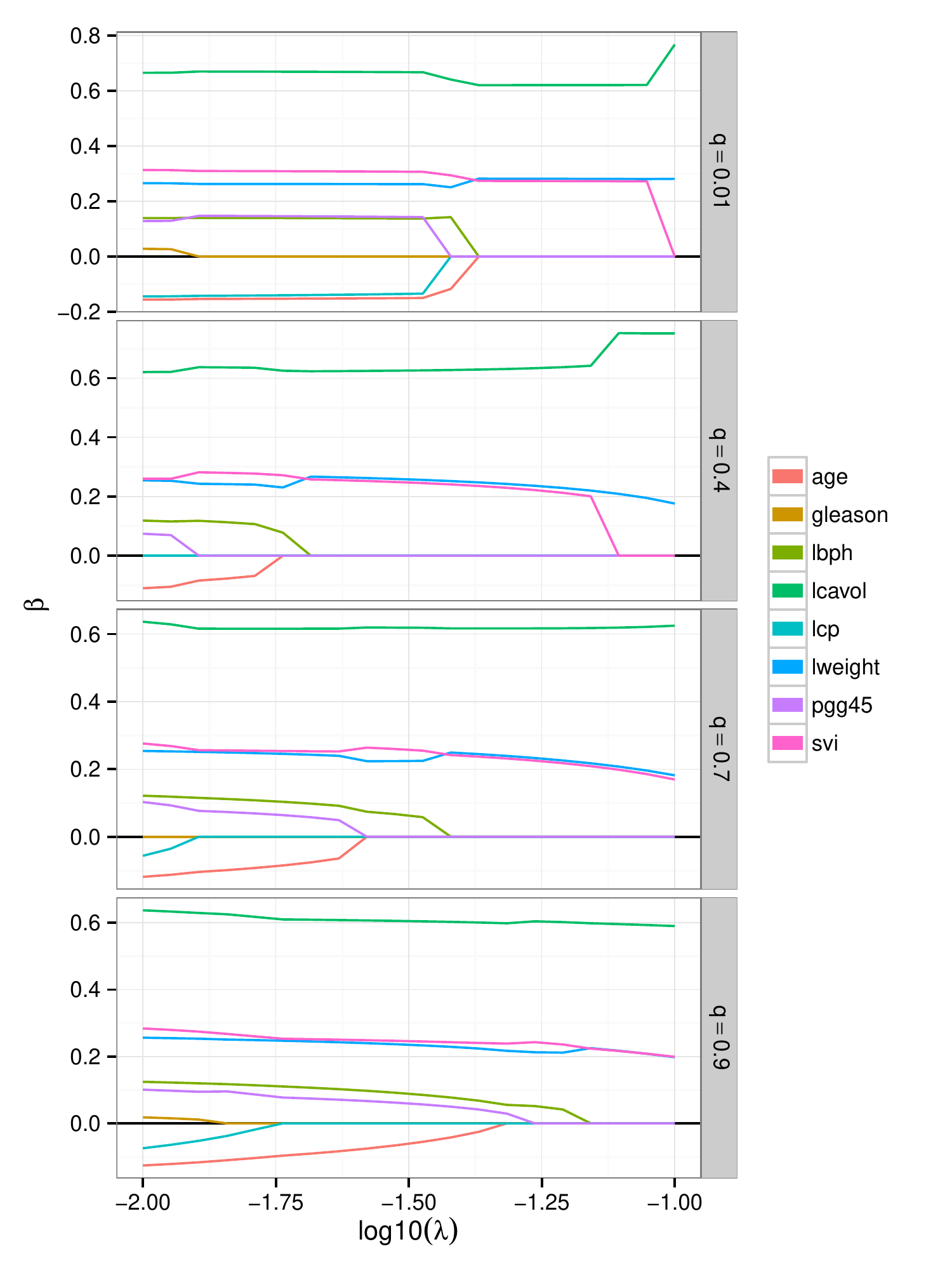}
    \caption{Proximal results for the prostate data example under the $L^q$-norm penalty.}
    \label{fig:prostate}
  \end{figure}

  A common regularized approach is to use
  lasso and elastic net, see \citet{tibshirani1996regression} and
  in \citet{zou2005regularization}, respectively.
  Alternatively, we fit the regularisation path using 
  $$
  \hat{x}^q_\lambda \defeq \argmin_x 
  \left\{ \frac{1}{2} \vnorm{y-Ax}^2 + \lambda \sum_{j=1}^p |x_i |^q \right\} \; .
  $$
  We can use the exact proximal operator for the $L^q$-norm and solve the
  harder non-convex problem.
  Figure~\ref{fig:prostate} shows the regularisation path.
  The major difference is, again, in the jumps to a sparse solution.

\section{Discussion}
\label{sec:discussion}

  Proximal algorithms are a widely applied approach to solving optimization problems
  that provide an extension of classical gradient descent
  methods and have properties that can be used to arrive at many different
  algorithmic implementations.  They are iterative shrinkage methods that extend traditional
  EM and MM algorithms--which are presently commonplace in statistics.
  \citet{beck2013weiszfeld} provide a historical perspective on iterative
  shrinkage algorithms by mainly focusing on the Weiszfeld algorithm
  \citep{weiszfeld1937point}.  The split Lagrangian methods described here were
  originally developed by \citet{hestenes1969multiplier} and
  \citet{rockafellar1973conjugate}.  More recently, there is work being done to
  extend the range of applicability of these methods outside of the class of
  convex functions to the broader class of functions satisfying the 
  Kurdyka-\L ojasiewicz inequality (see \cite{attouch2013convergence}).
   
  The purpose of our approach was to describe and apply the framework provided by
  proximal algorithms for constructing solutions to a large class of
  optimization problems in statistics.  These problems often involve composite
  functions that are representable by a sum of a linear or quadratic
  envelope together with a function that has a closed-form proximal operator
  that is easy to evaluate.  Numerous studies exist that demonstrate the
  efficacy and breadth of application of this approach.
  \citet{micchelli2013proximity,micchelli2011proximity} study proximal
  operators for composite operators for $L^2$-norm and $L^1$-norm/TV denoising models.
  \citet{argyriou2011efficient} describe numerical
  advantages of the proximal operator approach versus traditional fused lasso
  implementations. \citet{chen2013primal} provides a further class of fixed point
  algorithms that advance the proximal approach in the composite setting.

  Many MM block descent algorithms converge very slowly and there are a number
  of tools available to speed convergence. The most common approach involves
  Nesterov acceleration; see \citet{nesterov1983method} and
  \citet{beck2004conditional} who introduce a momentum term for gradient-descent
  algorithms applied to non-smooth composite problems.
  \citet{attouch2009convergence, noll2014convergence} provide further
  convergence properties for non-smooth functions. \citet{o2012adaptive} use
  adaptive restart to improve the convergence rate of accelerated gradient schemes.
  \citet{giselsson2014preconditioning} show how preconditioning can help with
  convergence for ill-conditioned problems.  \citet{meng2011accelerating} modify
  Nesterov's gradient method for strongly convex functions with Lipschitz
  continuous gradients.  \citet{allen2014novel} provide a simple interpretation
  of Nesterov's scheme as a two step algorithm with gradient-descent steps which
  yield proximal (forward) progress coupled with mirror-descent (backwards)
  steps with dual (backwards) progress.  By linearly coupling these two steps
  they improve convergence. \citet{giselsson2014preconditioning} show
  how preconditioning can help with convergence for ill-conditioned problems.

  There are a number of directions for future research on proximal methods in
  statistics, for example, exploring the use of Divide and Concur methods for
  mixed exponential family models, and the relationship between proximal
  splitting and variational Bayes methods in graphical models.  Another interesting
  area of research involves combining proximal steps with MCMC algorithms
  \citep{pereyra2013proximal}.  



\bibliographystyle{plainnat}
\bibliography{prox-methods}

\begin{thebibliography}{71}
\providecommand{\natexlab}[1]{#1}
\providecommand{\url}[1]{\texttt{#1}}
\expandafter\ifx\csname urlstyle\endcsname\relax
  \providecommand{\doi}[1]{doi: #1}\else
  \providecommand{\doi}{doi: \begingroup \urlstyle{rm}\Url}\fi

\bibitem[Allain et~al.(2006)Allain, Idier, and Goussard]{allain2006global}
Marc Allain, J{\'e}r{\^o}me Idier, and Yves Goussard.
\newblock On global and local convergence of half-quadratic algorithms.
\newblock \emph{Image Processing, IEEE Transactions on}, 15\penalty0
  (5):\penalty0 1130--1142, 2006.

\bibitem[Allen-Zhu and Orecchia(2014)]{allen2014novel}
Zeyuan Allen-Zhu and Lorenzo Orecchia.
\newblock A novel, simple interpretation of {N}esterov's accelerated method as
  a combination of gradient and mirror descent.
\newblock \emph{arXiv preprint arXiv:1407.1537}, 2014.

\bibitem[Argyriou et~al.(2011)Argyriou, Micchelli, Pontil, Shen, and
  Xu]{argyriou2011efficient}
Andreas Argyriou, Charles~A Micchelli, Massimiliano Pontil, Lixin Shen, and
  Yuesheng Xu.
\newblock Efficient first order methods for linear composite regularizers.
\newblock \emph{arXiv preprint arXiv:1104.1436}, 2011.

\bibitem[Attouch and Bolte(2009)]{attouch2009convergence}
Hedy Attouch and J{\'e}r{\^o}me Bolte.
\newblock On the convergence of the proximal algorithm for nonsmooth functions
  involving analytic features.
\newblock \emph{Mathematical Programming}, 116\penalty0 (1-2):\penalty0 5--16,
  2009.

\bibitem[Attouch et~al.(2010)Attouch, Bolte, Redont, and
  Soubeyran]{attouch2010proximal}
H{\'e}dy Attouch, J{\'e}r{\^o}me Bolte, Patrick Redont, and Antoine Soubeyran.
\newblock Proximal alternating minimization and projection methods for
  nonconvex problems: an approach based on the {K}urdyka-{L}ojasiewicz
  inequality.
\newblock \emph{Mathematics of Operations Research}, 35\penalty0 (2):\penalty0
  438--457, 2010.

\bibitem[Attouch et~al.(2013)Attouch, Bolte, and
  Svaiter]{attouch2013convergence}
Hedy Attouch, J{\'e}r{\^o}me Bolte, and Benar~Fux Svaiter.
\newblock Convergence of descent methods for semi-algebraic and tame problems:
  proximal algorithms, forward--backward splitting, and regularized
  gauss--seidel methods.
\newblock \emph{Mathematical Programming}, 137\penalty0 (1-2):\penalty0
  91--129, 2013.

\bibitem[Beck and Sabach(2013)]{beck2013weiszfeld}
Amir Beck and Shoham Sabach.
\newblock Weiszfeld's method: Old and new results.
\newblock \emph{Journal of Optimization Theory and Applications}, pages 1--40,
  2013.

\bibitem[Beck and Teboulle(2004)]{beck2004conditional}
Amir Beck and Marc Teboulle.
\newblock A conditional gradient method with linear rate of convergence for
  solving convex linear systems.
\newblock \emph{Mathematical Methods of Operations Research}, 59\penalty0
  (2):\penalty0 235--247, 2004.

\bibitem[Beck and Teboulle(2009)]{beck2009gradient}
Amir Beck and Marc Teboulle.
\newblock Gradient-based algorithms with applications to signal recovery.
\newblock \emph{Convex Optimization in Signal Processing and Communications},
  2009.

\bibitem[Beck and Teboulle(2014)]{beck2014fast}
Amir Beck and Marc Teboulle.
\newblock A fast dual proximal gradient algorithm for convex minimization and
  applications.
\newblock \emph{Operations Research Letters}, 42\penalty0 (1):\penalty0 1--6,
  2014.

\bibitem[Bertsekas(2011)]{bertsekas2011incremental}
Dimitri~P Bertsekas.
\newblock Incremental gradient, subgradient, and proximal methods for convex
  optimization: A survey.
\newblock \emph{Optimization for Machine Learning}, 2010:\penalty0 1--38, 2011.

\bibitem[Besag(1986)]{besag1986statistical}
Julian Besag.
\newblock On the statistical analysis of dirty pictures.
\newblock \emph{Journal of the Royal Statistical Society. Series B
  (Methodological)}, pages 259--302, 1986.

\bibitem[Bien et~al.(2013)Bien, Taylor, and Tibshirani]{bien2013lasso}
Jacob Bien, Jonathan Taylor, and Robert Tibshirani.
\newblock A lasso for hierarchical interactions.
\newblock \emph{The Annals of Statistics}, 41\penalty0 (3):\penalty0
  1111--1141, 2013.

\bibitem[Boyd and Vandenberghe(2009)]{boyd2009convex}
Stephen Boyd and Lieven Vandenberghe.
\newblock \emph{Convex optimization}.
\newblock Cambridge university press, 2009.

\bibitem[Boyd et~al.(2011)Boyd, Parikh, Chu, Peleato, and
  Eckstein]{boyd2011distributed}
Stephen Boyd, Neal Parikh, Eric Chu, Borja Peleato, and Jonathan Eckstein.
\newblock Distributed optimization and statistical learning via the alternating
  direction method of multipliers.
\newblock \emph{Foundations and Trends{\textregistered} in Machine Learning},
  3\penalty0 (1):\penalty0 1--122, 2011.

\bibitem[Bregman(1967)]{bregman1967relaxation}
Lev~M Bregman.
\newblock The relaxation method of finding the common point of convex sets and
  its application to the solution of problems in convex programming.
\newblock \emph{USSR Computational Mathematics and Mathematical Physics},
  7\penalty0 (3):\penalty0 200--217, 1967.

\bibitem[Cevher et~al.(2014)Cevher, Becker, and Schmidt]{cevher2014convex}
Volkan Cevher, Stephen Becker, and Mark Schmidt.
\newblock Convex optimization for big data: Scalable, randomized, and parallel
  algorithms for big data analytics.
\newblock \emph{Signal Processing Magazine, IEEE}, 31\penalty0 (5):\penalty0
  32--43, 2014.

\bibitem[Chambolle and Pock(2011)]{chambolle2011first}
Antonin Chambolle and Thomas Pock.
\newblock A first-order primal-dual algorithm for convex problems with
  applications to imaging.
\newblock \emph{Journal of Mathematical Imaging and Vision}, 40\penalty0
  (1):\penalty0 120--145, 2011.

\bibitem[Chaux et~al.(2007)Chaux, Combettes, Pesquet, and
  Wajs]{chaux2007variational}
Caroline Chaux, Patrick~L Combettes, Jean-Christophe Pesquet, and Val{\'e}rie~R
  Wajs.
\newblock A variational formulation for frame-based inverse problems.
\newblock \emph{Inverse Problems}, 23\penalty0 (4):\penalty0 1495, 2007.

\bibitem[Chen and Teboulle(1994)]{chen1994proximal}
Gong Chen and Marc Teboulle.
\newblock A proximal-based decomposition method for convex minimization
  problems.
\newblock \emph{Mathematical Programming}, 64\penalty0 (1-3):\penalty0 81--101,
  1994.

\bibitem[Chen et~al.(2013)Chen, Huang, and Zhang]{chen2013primal}
Peijun Chen, Jianguo Huang, and Xiaoqun Zhang.
\newblock A primal--dual fixed point algorithm for convex separable
  minimization with applications to image restoration.
\newblock \emph{Inverse Problems}, 29\penalty0 (2):\penalty0 025011, 2013.

\bibitem[Chouzenoux et~al.(2014)Chouzenoux, Pesquet, and
  Repetti]{chouzenoux2014variable}
Emilie Chouzenoux, Jean-Christophe Pesquet, and Audrey Repetti.
\newblock Variable metric forward--backward algorithm for minimizing the sum of
  a differentiable function and a convex function.
\newblock \emph{Journal of Optimization Theory and Applications}, 162\penalty0
  (1):\penalty0 107--132, 2014.

\bibitem[Combettes and Pesquet(2011)]{combettes2011proximal}
Patrick~L Combettes and Jean-Christophe Pesquet.
\newblock Proximal splitting methods in signal processing.
\newblock In \emph{Fixed-point algorithms for inverse problems in science and
  engineering}, pages 185--212. Springer, 2011.

\bibitem[Csiszar and Tusn{\'a}dy(1984)]{csisz1984information}
I~Csiszar and G{\'a}bor Tusn{\'a}dy.
\newblock Information geometry and alternating minimization procedures.
\newblock \emph{Statistics and decisions}, 1984.

\bibitem[Duckworth(2014)]{stronglyconvextable}
Daniel Duckworth.
\newblock The big table of convergence rates.
\newblock
  \url{https://github.com/duckworthd/duckworthd.github.com/blob/master/blog/big-table-of-convergence-rates.html},
  2014.

\bibitem[Efron(2011)]{efron2011tweedie}
Bradley Efron.
\newblock {T}weedie’s formula and selection bias.
\newblock \emph{Journal of the American Statistical Association}, 106\penalty0
  (496):\penalty0 1602--1614, 2011.

\bibitem[Esser et~al.(2010)Esser, Zhang, and Chan]{esser2010general}
Ernie Esser, Xiaoqun Zhang, and Tony~F Chan.
\newblock A general framework for a class of first order primal-dual algorithms
  for convex optimization in imaging science.
\newblock \emph{SIAM Journal on Imaging Sciences}, 3\penalty0 (4):\penalty0
  1015--1046, 2010.

\bibitem[Figueiredo and Nowak(2003)]{fig:nowak:2003}
M.A.T. Figueiredo and R.D. Nowak.
\newblock An {EM} algorithm for wavelet-based image restoration.
\newblock \emph{IEEE Transactions on Image Processing}, 12:\penalty0 906--16,
  2003.

\bibitem[Frankel et~al.(2014)Frankel, Garrigos, and
  Peypouquet]{frankelsplitting}
Pierre Frankel, Guillaume Garrigos, and Juan Peypouquet.
\newblock Splitting methods with variable metric for {KL} functions and general
  convergence rates.
\newblock 2014.

\bibitem[Geman and Reynolds(1992)]{geman1992constrained}
Donald Geman and George Reynolds.
\newblock Constrained restoration and the recovery of discontinuities.
\newblock \emph{IEEE Transactions on pattern analysis and machine
  intelligence}, 14\penalty0 (3):\penalty0 367--383, 1992.

\bibitem[Geman and Yang(1995)]{geman1995nonlinear}
Donald Geman and Chengda Yang.
\newblock Nonlinear image recovery with half-quadratic regularization.
\newblock \emph{Image Processing, IEEE Transactions on}, 4\penalty0
  (7):\penalty0 932--946, 1995.

\bibitem[Giselsson and Boyd(2014)]{giselsson2014preconditioning}
Pontus Giselsson and Stephen Boyd.
\newblock Preconditioning in fast dual gradient methods.
\newblock In \emph{Proceedings of the 53rd Conference on Decision and Control},
  2014.

\bibitem[Gravel and Elser(2008)]{gravel2008divide}
Simon Gravel and Veit Elser.
\newblock Divide and concur: A general approach to constraint satisfaction.
\newblock \emph{Physical Review E}, 78\penalty0 (3):\penalty0 036706, 2008.

\bibitem[Green(1990)]{green1990use}
Peter~J Green.
\newblock On use of the {EM} for penalized likelihood estimation.
\newblock \emph{Journal of the Royal Statistical Society. Series B
  (Methodological)}, pages 443--452, 1990.

\bibitem[{Green} et~al.(2015){Green}, {{\L}atuszy{\'n}ski}, {Pereyra}, and
  {Robert}]{2015arXiv150201148G}
Peter~J. {Green}, K.~{{\L}atuszy{\'n}ski}, M.~{Pereyra}, and C.~P. {Robert}.
\newblock {Bayesian computation: a perspective on the current state, and
  sampling backwards and forwards}.
\newblock \emph{ArXiv e-prints}, February 2015.

\bibitem[Hastie et~al.(2009)Hastie, Tibshirani, Friedman, Hastie, Friedman, and
  Tibshirani]{hastie2009elements}
Trevor Hastie, Robert Tibshirani, Jerome Friedman, T~Hastie, J~Friedman, and
  R~Tibshirani.
\newblock \emph{The elements of statistical learning}, volume~2.
\newblock Springer, 2009.

\bibitem[Hestenes(1969)]{hestenes1969multiplier}
Magnus~R Hestenes.
\newblock Multiplier and gradient methods.
\newblock \emph{Journal of optimization theory and applications}, 4\penalty0
  (5):\penalty0 303--320, 1969.

\bibitem[Hu et~al.()Hu, Li, and Yang]{huproximal}
YH~Hu, C~Li, and XQ~Yang.
\newblock Proximal gradient algorithm for group sparse optimization.

\bibitem[Komodakis and Pesquet(2014)]{komodakis2014playing}
Nikos Komodakis and Jean-Christophe Pesquet.
\newblock Playing with duality: An overview of recent primal-dual approaches
  for solving large-scale optimization problems.
\newblock \emph{arXiv preprint arXiv:1406.5429}, 2014.

\bibitem[Magn{\'u}sson et~al.(2014)Magn{\'u}sson, Weeraddana, Rabbat, and
  Fischione]{magnusson2014convergence}
Sindri Magn{\'u}sson, Pradeep~Chathuranga Weeraddana, Michael~G Rabbat, and
  Carlo Fischione.
\newblock On the convergence of alternating direction lagrangian methods for
  nonconvex structured optimization problems.
\newblock \emph{arXiv preprint arXiv:1409.8033}, 2014.

\bibitem[Marjanovic and Solo(2013)]{marjanovic2013exact}
Goran Marjanovic and Victor Solo.
\newblock On exact $\ell^q$ denoising.
\newblock In \emph{Acoustics, Speech and Signal Processing (ICASSP), 2013 IEEE
  International Conference on}, pages 6068--6072. IEEE, 2013.

\bibitem[Martinet(1970)]{martinet1970breve}
Bernard Martinet.
\newblock Br{\`e}ve communication. r{\'e}gularisation d'in{\'e}quations
  variationnelles par approximations successives.
\newblock \emph{ESAIM: Mathematical Modelling and Numerical
  Analysis-Mod{\'e}lisation Math{\'e}matique et Analyse Num{\'e}rique},
  4\penalty0 (R3):\penalty0 154--158, 1970.

\bibitem[Meng and Chen(2011)]{meng2011accelerating}
Xiangrui Meng and Hao Chen.
\newblock Accelerating {N}esterov's method for strongly convex functions with
  lipschitz gradient.
\newblock \emph{arXiv preprint arXiv:1109.6058}, 2011.

\bibitem[Micchelli et~al.(2011)Micchelli, Shen, and Xu]{micchelli2011proximity}
Charles~A Micchelli, Lixin Shen, and Yuesheng Xu.
\newblock Proximity algorithms for image models: denoising.
\newblock \emph{Inverse Problems}, 27\penalty0 (4):\penalty0 045009, 2011.

\bibitem[Micchelli et~al.(2013)Micchelli, Shen, Xu, and
  Zeng]{micchelli2013proximity}
Charles~A Micchelli, Lixin Shen, Yuesheng Xu, and Xueying Zeng.
\newblock Proximity algorithms for the {L}1/{TV} image denoising model.
\newblock \emph{Advances in Computational Mathematics}, 38\penalty0
  (2):\penalty0 401--426, 2013.

\bibitem[Nesterov(1983)]{nesterov1983method}
Yurii Nesterov.
\newblock A method of solving a convex programming problem with convergence
  rate ${O}(1/k^2)$.
\newblock In \emph{Soviet Mathematics Doklady}, volume~27, pages 372--376,
  1983.

\bibitem[Nikolova and Ng(2005)]{nikolova2005analysis}
Mila Nikolova and Michael~K Ng.
\newblock Analysis of half-quadratic minimization methods for signal and image
  recovery.
\newblock \emph{SIAM Journal on Scientific computing}, 27\penalty0
  (3):\penalty0 937--966, 2005.

\bibitem[Noll(2014)]{noll2014convergence}
Dominikus Noll.
\newblock Convergence of non-smooth descent methods using the
  {K}urdyka--{\l}ojasiewicz inequality.
\newblock \emph{Journal of Optimization Theory and Applications}, 160\penalty0
  (2):\penalty0 553--572, 2014.

\bibitem[O'Donoghue and Candes(2012)]{o2012adaptive}
Brendan O'Donoghue and Emmanuel Candes.
\newblock Adaptive restart for accelerated gradient schemes.
\newblock \emph{Foundations of Computational Mathematics}, pages 1--18, 2012.

\bibitem[Palmer et~al.(2005)Palmer, Kreutz-Delgado, Rao, and
  Wipf]{palmer2005variational}
Jason Palmer, Kenneth Kreutz-Delgado, Bhaskar~D Rao, and David~P Wipf.
\newblock Variational {EM} algorithms for non-gaussian latent variable models.
\newblock In \emph{Advances in neural information processing systems}, pages
  1059--1066, 2005.

\bibitem[Parikh and Boyd(2013)]{parikh2013proximal}
Neal Parikh and Stephen Boyd.
\newblock Proximal algorithms.
\newblock \emph{Foundations and Trends in Optimization}, 1\penalty0
  (3):\penalty0 123--231, 2013.

\bibitem[Patrinos and Bemporad(2013)]{patrinos2013proximal}
Panagiotis Patrinos and Alberto Bemporad.
\newblock Proximal newton methods for convex composite optimization.
\newblock In \emph{Decision and Control (CDC), 2013 IEEE 52nd Annual Conference
  on}, pages 2358--2363. IEEE, 2013.

\bibitem[Patrinos et~al.(2014)Patrinos, Stella, and
  Bemporad]{patrinos2014douglas}
Panagiotis Patrinos, Lorenzo Stella, and Alberto Bemporad.
\newblock Douglas-rachford splitting: complexity estimates and accelerated
  variants.
\newblock \emph{arXiv preprint arXiv:1407.6723}, 2014.

\bibitem[Pereyra(2013)]{pereyra2013proximal}
Marcelo Pereyra.
\newblock Proximal markov chain monte carlo algorithms.
\newblock \emph{arXiv preprint arXiv:1306.0187}, 2013.

\bibitem[Polson and Scott(2012)]{Polson:Scott:2010b}
Nicholas~G. Polson and James~G. Scott.
\newblock Local shrinkage rules, {L}\'evy processes, and regularized
  regression.
\newblock \emph{Journal of the Royal Statistical Society (Series B)},
  74\penalty0 (2):\penalty0 287--311, 2012.

\bibitem[Polson and Scott(2014)]{polson2014mixtures}
Nicholas~G Polson and James~G Scott.
\newblock Mixtures, envelopes, and hierarchical duality.
\newblock \emph{arXiv preprint arXiv:1406.0177}, 2014.

\bibitem[Quiroz and Oliveira(2009)]{quiroz2009proximal}
EA~Papa Quiroz and P~Roberto Oliveira.
\newblock Proximal point methods for quasiconvex and convex functions with
  {B}regman distances on hadamard manifolds.
\newblock \emph{J. Convex Anal}, 16\penalty0 (1):\penalty0 46--69, 2009.

\bibitem[Robbins(1964)]{robbins1964empirical}
Herbert Robbins.
\newblock The empirical bayes approach to statistical decision problems.
\newblock \emph{The Annals of Mathematical Statistics}, pages 1--20, 1964.

\bibitem[Rockafellar and Wets(1998)]{rockafellar:wets:1998}
R.~Tyrell Rockafellar and R.~J-B Wets.
\newblock \emph{Variational Analysis}.
\newblock Springer, 1998.

\bibitem[Rockafellar(1973)]{rockafellar1973conjugate}
R~Tyrrell Rockafellar.
\newblock Conjugate duality and optimization.
\newblock Technical report, DTIC Document, 1973.

\bibitem[Rockafellar(1976)]{rockafellar1976monotone}
R~Tyrrell Rockafellar.
\newblock Monotone operators and the proximal point algorithm.
\newblock \emph{SIAM Journal on Control and Optimization}, 14\penalty0
  (5):\penalty0 877--898, 1976.

\bibitem[Rudin et~al.(1992)Rudin, Osher, and Faterni]{rudin:osher:faterni:1992}
L.~Rudin, S.~Osher, and E.~Faterni.
\newblock Nonlinear total variation based noise removal algorithms.
\newblock \emph{Phys.~D}, 60\penalty0 (259--68), 1992.

\bibitem[Tansey et~al.(2014)Tansey, Koyejo, Poldrack, and
  Scott]{tansey:etal:2014}
Wesley Tansey, Oluwasanmi Koyejo, Russell~A. Poldrack, and James~G. Scott.
\newblock False discovery rate smoothing.
\newblock Technical report, University of Texas at Austin, 2014.

\bibitem[Tibshirani et~al.(2005)Tibshirani, Saunders, Rosset, Zhu, and
  Knight]{tibs:fusedlasso:2005}
R.~Tibshirani, M.~Saunders, S.~Rosset, J.~Zhu, and K.~Knight.
\newblock Sparsity and smoothness via the fused lasso.
\newblock \emph{Journal of the Royal Statistical Society (Series B)},
  67:\penalty0 91--108, 2005.

\bibitem[Tibshirani(2014)]{tibs:2014a}
R.J. Tibshirani.
\newblock Adaptive piecewise polynomial estimation via trend filtering.
\newblock \emph{Annals of Statistics}, 42\penalty0 (1):\penalty0 285--323,
  2014.

\bibitem[Tibshirani(1996)]{tibshirani1996regression}
Robert Tibshirani.
\newblock Regression shrinkage and selection via the lasso.
\newblock \emph{Journal of the Royal Statistical Society. Series B
  (Methodological)}, pages 267--288, 1996.

\bibitem[Von~Neumann(1951)]{von1951functional}
John Von~Neumann.
\newblock \emph{Functional operators: The geometry of orthogonal spaces}.
\newblock Princeton University Press, 1951.

\bibitem[Weiszfeld(1937)]{weiszfeld1937point}
Endre Weiszfeld.
\newblock Sur le point pour lequel la somme des distances de n points
  donn{\'e}s est minimum.
\newblock \emph{Tohoku Math. J}, 43\penalty0 (355-386):\penalty0 2, 1937.

\bibitem[Witten et~al.(2009)Witten, Tobshirani, and
  Hastie]{witten:tibs:hastie:2009}
Daniella~M. Witten, Robert Tobshirani, and Trevor Hastie.
\newblock A penalized matrix decomposition, with applications to sparse
  principal components and canonical correlation analysis.
\newblock \emph{Biostatistics}, 10\penalty0 (3):\penalty0 515--34, 2009.

\bibitem[Zhang et~al.(2010)Zhang, Saha, and Vishwanathan]{zhang2010regularized}
Xinhua Zhang, Ankan Saha, and SVN Vishwanathan.
\newblock Regularized risk minimization by {N}esterov's accelerated gradient
  methods: {A}lgorithmic extensions and empirical studies.
\newblock \emph{arXiv preprint arXiv:1011.0472}, 2010.

\bibitem[Zou and Hastie(2005)]{zou2005regularization}
Hui Zou and Trevor Hastie.
\newblock Regularization and variable selection via the elastic net.
\newblock \emph{Journal of the Royal Statistical Society: Series B (Statistical
  Methodology)}, 67\penalty0 (2):\penalty0 301--320, 2005.

\end{thebibliography}

\begin{appendix}
  
%
%
\newcommand{\heading}[1]{{\centering#1}}

\newcolumntype{R}{>{\raggedright\arraybackslash}X}%
\def\arraystretch{1.6}

\begin{table}[l]
{ \footnotesize
  \begin{tabularx}{1.1\textwidth}{@{}|
      p{3cm}|
      R|
      R|
    @{}}

  \hline
  \heading{Type}
  & \heading{$\phi(x)$} 
  & \heading{$\prox_{\gamma\phi}(y)$} 
  \\
  \hline
    Laplace 
    & $\omega \|x\|$ 
  & 
    $\sgn(x)\max(\|x\|-\omega,0)$ 
  \\ 
    Gaussian
  & $\tau \|x\|^2$
  & $x/(2\tau + 1)$
  \\
    Group-sparse, $\ell_p$ 
  & $\kappa \left\|x\right\|^{p}$ 
  & 
    $\sgn(x) \rho$, 
    \newline 
    $\rho$ s.t. $\rho + p \kappa \rho^{p-1}=\|x\|$
  \\
    \vdots
  & $p=4/3$
  & 
    $x + \frac{4\kappa}{3 2^{1/3}}\left((\chi - x)^{1/3}-(\chi+x)^{1/3}\right)$ 
    \newline
    $\chi = \sqrt{x^2 + 256 \kappa^3/729}$
  \\
    \vdots
  & $p=3/2$
  & 
    $x + 9 \kappa^2 \sgn(x)\left(1-\sqrt{1+16|x|/(9\kappa^2)}\right)/8$
  \\
    \vdots
  & $p=3$
  & 
    $\sgn(x)\left(\sqrt{1+12\kappa|x|}-1\right)/(6\kappa)$
  \\
    \vdots
  & $p=4$
  & 
    $\left(\frac{\chi+x}{8\kappa}\right)^{1/3}-\left(\frac{\chi-x}{8\kappa}\right)^{1/3}$
    \newline
    $\chi=\sqrt{x^2+1/(27\kappa)}$
  \\
    Gamma, Chi 
  & $-\kappa \ln x + \omega x$ 
  & 
    $\frac{1}{2} \left(x-\omega+\sqrt{(x- \omega)^2 + 4 \kappa}\right)$ 
  \\
    Double-Pareto
  & 
    $\gamma \log(1+|x|/a)$
  & 
    $\frac{\sgn(x)}{2} \left\{|x| - a + \sqrt{ (a- |x|)^2 + 4 d(x) } \right\}$,
    \newline
    $d(x) = (a|x| - \gamma)_+$
  \\ 
    Huber dist.
  & 
    $\begin{cases}
      \tau x^2 & |x| \le \omega/\sqrt{2\tau}\\
      \omega \sqrt{2\tau}|x|-\omega^2/2 & \text{otherwise}
    \end{cases}$ 
    \newline
    $\omega, \tau \in (0,+\infty)$
  & 
    $\begin{cases}
      \frac{x}{2\tau+1} & |x| \le \omega(2\tau+1)/\sqrt{2\tau}\\
      x-\omega\sqrt{2\tau}\sgn(x) & |x| > \omega(2\tau+1)/\sqrt{2\tau}
    \end{cases}$
  \\
    Max-entropy dist.
  & 
    $\omega|x|+\tau|x|^2+\kappa|x|^p$ 
    \newline
    $2 \ne p \in (1,+\infty)$, 
    \newline
    $\omega,\tau,\kappa \in (0,+\infty)$
  & 
    $\sgn(x) \underset{\kappa|\cdot|^p/(2\tau+1)}{\prox}\left(\frac{1}{2\tau+1}\max(|x|-\omega,0)\right)$
  \\
    Smoothed-laplace dist.
  & 
    $\omega |x| - \ln(1+\omega|x|)$
  & 
    $\sgn(x)\frac{\omega|x|-\omega^2-1+\sqrt{\left|\omega|x|-\omega^2-1\right|^2+4\omega|x|}}{2\omega}$
  \\
    Exponential dist.
  & 
    $\begin{cases}
      \omega x & x \ge 0\\
      +\infty & x < 0
    \end{cases}$
  & 
    $\begin{cases}
      x-\omega & x \ge \omega\\
      0 & x < \omega
    \end{cases}$
  \\
    Uniform dist.
  & 
    $\begin{cases}
      -\omega & x < -\omega\\
      x & |x| \le \omega\\
      \omega & x > \omega
    \end{cases}$
  & 
    $\begin{cases}
      x-\omega & x \ge \omega\\
      0 & x < \omega
    \end{cases}$
  \\
    Triangular dist.
  & 
   $\begin{cases}
      -\ln(x-\omega) + \ln(-\omega) & x \in (\omega,0)\\
      -\ln(\hat{\omega}-x) + \ln(\hat{\omega}) & x \in (0,\hat{\omega})\\
      +\infty & \text{otherwise}
    \end{cases}$ 
    \newline
    $\omega \in (-\infty,0]$,
    $\hat{\omega} \in (0,\infty)$
  & 
   $\begin{cases}
      \frac{x+\omega+\sqrt{|x-\omega|^2+4}}{2} & x < 1/\omega\\
      \frac{x+\hat{\omega}-\sqrt{|x-\hat{\omega}|^2+4}}{2} & x > 1/\hat{\omega}
    \end{cases}$
  \\
    Weibull dist.
  & 
   $\begin{cases}
      -\kappa \ln x + \omega x^p & x >0 \\
      +\infty & x \le 0
    \end{cases}$ 
    \newline
    $p \in (1,+\infty)$
    $\omega, \kappa \in (-\infty,0]$
  & 
    $\pi$ s.t. $p \omega \pi^p + \pi^2 - x \pi = \kappa$
  \\
  GIG dist.
  & 
   $\begin{cases}
      -\kappa \ln x + \omega x + \rho/x & x >0 \\
      +\infty & x \le 0
    \end{cases}$ 
    \newline
    $\omega,\kappa,\rho \in (-\infty,0]$
  & 
    $\pi$ s.t. $\pi^3+(\omega-x)\pi^2-\kappa \pi = \rho$
  \\

  \hline
\end{tabularx}
}
\caption{Sources: \citep{chaux2007variational} \citep{huproximal}}
\label{tab:prox}
\end{table}



\begin{table}[l]
{ \footnotesize
  \begin{tabularx}{1.1\textwidth}{@{}|p{5cm}||R|R|@{}}
  \hline
  \heading{Penalty}
  & \multicolumn{2}{c|}{\heading{Minimizer}}
  \\  
  \hline
    \heading{$\phi(t) = \min_s \left\{Q(t,s) + \psi(s)\right\}$} 
  & \heading{$Q(t,s) = \frac{1}{2} t^2 s $} 
  & \heading{$Q(t,s) = (t-s)^2 $} 
  \\
  \hline
    $|t|^\alpha$, 
    $\alpha \in (1,2]$ 
  & $\alpha |t|^{\alpha-2}$
  &  
  \\
    $\sqrt{\alpha + t^2}$ 
  & $\frac{1}{\sqrt{\alpha + t^2}}$
  & $c t - \frac{t}{\sqrt{\alpha + t^2}}$
  \\
    $\frac{|t|}{\alpha} - \log\left(1+\frac{|t|}{\alpha}\right)$ 
  & $\frac{1}{\alpha(\alpha + |t|)}$
  & $c t - \frac{t}{\alpha(\alpha + |t|)}$
  \\
    $\begin{cases}
      \frac{t^2}{2} & |t| \le \alpha \\
      \alpha |t| - \frac{\alpha^2}{2} & |t| > \alpha
     \end{cases}$ 
  & 
    $\begin{cases}
      1 & |t| \le \alpha \\
      \frac{\alpha}{|t|} & |t| > \alpha
     \end{cases}$ 
  & 
    $\begin{cases}
      (c-1)t & |t| \le \alpha \\
      c t - \alpha \sgn(t) & |t| > \alpha
     \end{cases}$ 
  \\
    $\log(\cosh(\alpha t))$ 
  & $\alpha \frac{\tanh(\alpha t)}{t}$
  & $c t - \alpha \tanh(\alpha t)$
  \\
    $-\frac{1}{1+|\bx|}$ 
  & $\begin{cases} 
        -2 & \text{for}\: t = 0 \\
        \frac{\operatorname{sgn}{\left (t \right )}}{t \left(\left\lvert{t}\right\rvert + 1\right)^{2}} & \text{otherwise} 
      \end{cases}$
  & $c t - \frac{\operatorname{sgn}{\left (t \right )}}{\left(\left\lvert{t}\right\rvert + 1\right)^{2}}$
  \\
  $-\frac{1}{1+\sqrt{\bx}}$ 
  & 
  $\begin{cases} 
    -\infty & \text{for}\: t = 0 \\
    \frac{1}{2 t^{\frac{3}{2}} \left(\sqrt{t} + 1\right)^{2}} & \text{otherwise} 
  \end{cases}$
  & $c t - \frac{1}{2 \sqrt{t} \left(\sqrt{t} + 1\right)^{2}}$
  \\
  \hline
\end{tabularx}
}
\caption[...]{
  Minimizers for the multiplicative form are 
  \texorpdfstring{
    $\sigma(t) =\begin{cases}\phi^{''}(0^+) & \text{if } t = 0,\\
    \phi^{'}(t)/t & \text{if } t \neq 0
    \end{cases}$
  }{math} ,
  and for additive form
  \texorpdfstring{$\sigma(t) = c t - \phi^{'}(t)$}{math}.
  See \citep{nikolova2005analysis}.
}
\label{tab:hq}
\end{table}



\begin{table}[l]
{ \footnotesize
  \begin{tabularx}{1.1\textwidth}{@{}|R|R|R|R|@{}}
  \hline
  & \multicolumn{2}{c|}{\heading{Error Rate}}
  & 
  \\  
  \hline
  \heading{Algorithm}
  & Convex
  & Strongly Convex
  & \heading{Per-Iteration Cost}
  \\
  \hline
  Accelerated Gradient Descent
  &
  $O(1/\sqrt{\epsilon})$
  & 
  $O(\log(1/\epsilon))$
  & 
  $O(n)$
  \\
  \hline
  Proximal Gradient Descent
  &
  $O(1/\epsilon)$
  & 
  $O(\log(1/\epsilon))$
  & 
  $O(n)$
  \\
  \hline
  Accelerated Proximal Gradient Descent
  &
  $O(1/\sqrt{\epsilon})$
  & 
  $O(\log(1/\epsilon))$
  & 
  $O(n)$
  \\
  \hline
  ADMM
  &
  $O(1/\epsilon)$
  & 
  $O(\log(1/\epsilon))$
  & 
  $O(n)$
  \\
  \hline
  Frank-Wolfe / Conditional Gradient Algorithm
  &
  $O(1/\epsilon)$
  & 
  $O(1/\sqrt{\epsilon})$
  & 
  $O(n)$
  \\
  \hline
  Newton's Method
  &
  & 
  $O(\log\log(1/\epsilon))$
  & 
  $O(n^3)$
  \\
  \hline
  Conjugate Gradient Descent
  &
  & 
  $O(n)$
  & 
  $O(n^2)$
  \\
  \hline
  L-BFGS
  &
  & 
  Between 
  $O(\log(1/\epsilon))$ and
  $O(\log\log(1/\epsilon))$
  & 
  $O(n^2)$
  \\
  \hline
\end{tabularx}
}
\caption{
  See \citep{stronglyconvextable}.
}
\label{tab:rates}
\end{table}

  \section{Convergence}
  \label{app:convergence}
     
    We now establish convergence results for the forward-backward proximal solution to
    \eqref{eq:quad_objective} given in \eqref{eqn:fixedpointproxgrad}
    $$
    x^{\star} = \prox_{\phi/\lambda }\{ x - \nabla l(x)/\lambda  \} \, ,
    $$
    when $l$ and $\phi$ are lower semi-continuous and $\nabla l$ is Lipschitz
    continuous.  We also assume that $\prox_{\phi/\lambda}$ is non-empty and can
    be evaluated independently in each component of $y$.


    Recalling the translation property of proximal operators
    stated in \ref{eq:prox_translate}, we can say
    \begin{align*}
      x^{\star} &= \prox_{\phi/\lambda}\left( x - \nabla l(x)/\lambda \right) 
                = \prox_{(\phi(z) + \lambda {\nabla l(z)}^T z)/\lambda }\left( x \right) \\ 
        &=\argmin_z \left\{\phi(z) + {\nabla l(z)}^T (z - x) 
          + \frac{\lambda}{2} \|x-z\|^2  \right\}
    \end{align*}
    By the proximal operator's minimizing properties, its solution $x^{\star}$
    satisfies
    $$
     \phi(x^{\star}) 
       + {\nabla l(x^{\star})}^T (x^{\star} - x) 
       + \frac{\lambda}{2} \|x-x^{\star}\|^2 
     \leq \phi(x)   
    $$
    providing a sort of quadratic minorizer for $F(w)$ in the form of
    $$
    l(w) + \phi(x^{\star}) 
       + {\nabla l(x^{\star})}^T (x^{\star} - w) 
       + \frac{\lambda}{2} \|w-x^{\star}\|^2 
       \leq l(w) + \phi(w) \equiv F(w)
    $$
    The Lipschitz continuity of $\nabla l(x)$, i.e.
    $$
    l(x) \leq l(w) + {\nabla l(w)}^T (x - w) + \frac{\gamma}{2} \|x-w\|^2 \;,
    $$                                
    also gives us a quadratic majorizer
    \begin{align*}
      F(x) \equiv l(x) + \phi(x) &\leq l(w) + {\nabla l(w)}^T (w - x) 
        + \frac{\gamma}{2} \|x-x^{\star}\|^2 
    \end{align*}
    which, when evaluated at $x = x^{\star}$ and combined with our minorizer
    yields
    \begin{align*}
      (\lambda - \gamma) \half \|x^{\star} - w\|^2 
        &\leq F(w) - F(x^{\star})
    \end{align*}
    Thus, if we want to ensure that the objective value will decrease in this
    procedure, we need to fix $\lambda \geq \gamma$.  Furthermore, functional
    characteristics of $l$ and $\phi$, such as convexity, can improve the
    bounds in the steps above and guarantee good--or optimal--decreases in
    $F(w) - F(x^{\star})$.

    Finally, when we compound up the errors we obtain a $O(1/k)$ convergence
    bound.  This can be improved by adding a momentum term to $y$ that includes the
    first derivative information.
    
    These arguments can be extended to Bregman divergences by way of the general
    law of cosines inequality
    $$
    D(x,w) = D(x,z) +D(w,z) + (\nabla l(z) - \nabla l(w) )^T (x-w) \;,
    $$
    so that $ D(x,w) \geq D(x, P(w)) + D( P(w),w) $ where $ P(w) = \argmin_v D(v,w)$.
 
 \section{Nesterov Acceleration}
  \label{app:acceleration}

    A powerful addition is Nesterov acceleration. Consider a
    convex combination, with parameter $\theta$,  of upper bounds for the proximal
    operator inequality $ z=x $ and $ z= x^\star $.
    We are free to choose variables $z= \theta x + (1 - \theta) x^+ $ and $w$. 
    If $\phi$ is convex, 
    $\phi(\theta x + (1-\theta) x^+) \leq \theta \phi(x) + (1-\theta) \phi(x^+)$,
    then we have 
    \begin{align*}
      F( x^+ ) & - F^\star - (1 - \theta ) (F(x) - F^\star )\\
               & = F( x^+ ) - \theta F^\star - (1-\theta) F(x)\\
               & \leq \lambda ( x^+ - w)^T (\theta x^\star + (1-\theta) x - x^+ ) 
      + \frac{\lambda}{2} \vnorm{x^+-w}^2\\
               & =  \frac{\lambda}{2} \left ( \vnorm{w - (1 -\theta) x - \theta x^\star}^2 
      - \vnorm{x^+ - (1 -\theta) x - \theta x^\star}^2 \right ) \\
               & = \frac{\theta^2 \lambda}{2} \left ( \vnorm{u - x^\star}^2 
      - \vnorm{u^+ - x^\star}^2 \right )
    \end{align*}
    Where $w$ is given in terms of the intermediate steps 
    \begin{align*}
      \theta u & = w - ( 1 - \theta ) x \\
      \theta u^+ & = x^+ - (1 - \theta)x
    \end{align*}
    Introducing a sequence $\theta_t$ with iteration subscript, $t$. 
    The second identity, $ \theta u = x - (1- \theta) x^- $, then yields an
    update for $w$ as the current state $x$ plus a momentum term, 
    depending on the direction $(x-x^-)$,
    namely 
    $$
    w = (1- \theta_t) x + \theta_t u = x - \theta_{t-1} (1 - \theta_t ) ( x - x^- )
    $$
   
  \section{Quasi-convex Convergence}
  \label{app:quasi-convex}

    Consider an optimisation problem
    $ \min_{ x \in \mathcal{X}} l(x) $ where $l$ is quasi-convex, continuous and has
    non-empty set of finite global minima.  Let $x^t$ be generated by the
    proximal point algorithm 
    $$
    x^t \in {\rm arg min} \left\{ l(x) + \frac{\lambda_t}{2} \|x- x^t \|^2 \right\} \;.
    $$
    \citet{quiroz2009proximal} show that these iterates converge to the global
    minima, although the proximal operator at each step may be set-valued--due to
    the non-convexity of $l$.
    A function $l$ is quasi-convex when
    $$
    l(\theta x + (1-\theta)z) \leq \max(l(x),l(z)) \;,
    $$
    which accounts for a number of non-convex functions like $|x|^q$, when $ 0 < q <1$, 
    and functions involving appropriate ranges of $\log(x)$ and $\tanh(x)$. 
    In this setting, using the level-sets generated by the sequence, i.e. 
    $ U = \left\{ x \in \text{dom} (l) : l(x) \leq \inf_t l(x^t) \right\} $, one finds
    that $U$ is a non-empty closed convex set and that
    $x^t$ is a Fej\'er sequence of finite length, $ \sum_t \|x^{t+1} - x^t \| < \infty $,
    and that it converges to a critical point of $l$ as long as 
    $ \min \left\{ l(x) : x\in \Re^d \right\} $ is nonempty.

  \section{Non-convex: Kurdyka-\L ojasiewicz (KL)}
  \label{app:kl}

    A locally Lipschitz function $l : \Re^d \rightarrow \Re $ satisfies KL at 
    $x^\star \in \Re^d $ if and only if $ \exists \eta \in (0, \infty)$ 
    and a neighbourhood $U$ of $ x^\star$ and a concave 
    $ \kappa : [0,\eta] \rightarrow [0,\infty)$ with 
      $\kappa(0)=0$, $\kappa \in C^1, \kappa^\prime > 0 $ on $(0,\eta)$ 
    and for every $x \in U$ with $ l(x^\star) < l(x) < l(x^\star) + \eta $ we have
    $$
    \kappa^\prime \left\{ l(x) - l(x^\star) \right\} 
      \operatorname{dist}\left(0, \partial l(x) \right ) \geq 1 
    $$
    where 
    $\operatorname{dist}(0, A) \equiv \underset{ x \in A}{\sup} \| x \|^2 $.

    The KL condition guarantees summability and therefore a finite length of the
    discrete subgradient trajectory.
    Using the KL properties of a function, one can show convergence for alternating
    minimisation algorithms for problems like
    $$
    \min_{x,z} L(x,z) \defeq l(x) + Q(x,z) + \phi(z) \;,
    $$
    where $ \nabla Q$ is Lipschitz continuous 
    (see \citet{attouch2010proximal, attouch2013convergence}).
    A typical application involves solving 
    $ \min_{x \in \Re^d} \left\{ l(x)+\phi(x) \right\} $ via the augmented Lagrangian
    $$ 
    L(x,z) =l(x)+\phi(z)+\lambda^\top(x-z) + \frac{\rho}{2} \|x-z \|^2
    $$ 
    where $\rho$ is a relaxation parameter.

    A useful class of functions that satisfy KL as ones that possess uniform convexity
    $$
    l(y) \geq l(x) + u^\top (z-x) + K \|z-x \|^p, 
    \text{ where } p \geq 1 \; \; , \forall u \in \partial l(x) \; .
    $$
    Then $l$ satisfies KL on $ \text{dom} (l) $ for 
    $\kappa (s) = p K^{- \frac{1}{p}} s^{ \frac{1}{p}}$.

    For explicit convergence rates in the KL setting, see \citep{frankelsplitting}.
\end{appendix}

\end{document}